\documentclass[a4paper]{cas-sc}

\usepackage[authoryear]{natbib}
\usepackage{lipsum}
\usepackage{algorithm}
\usepackage{algpseudocode}
\usepackage{amsmath}
\usepackage{amssymb}
\usepackage{float}
\usepackage{cancel}
\usepackage{enumitem}
\usepackage{mathtools}
\usepackage{multirow}
\usepackage{dsfont}
\setcitestyle{square}
\usepackage{subfigure}
\usepackage{hhline} 
\usepackage{highlight}
\usepackage{cleveref}
\usepackage[inkscapelatex=false]{svg}
\usepackage{soul}
\usepackage{bm}
\usepackage[singlelinecheck=false]{caption}

\usepackage[flushleft]{threeparttable}
\usepackage{acronym}
\usepackage{framed}
\usepackage{nomencl} 
\makenomenclature

\algnewcommand\algorithmicforeach{\textbf{for each:}}
\algnewcommand\ForEach{\item[ \algorithmicforeach]}

\DeclareUnicodeCharacter{0398}{\ensuremath{\Theta}}
\DeclareUnicodeCharacter{0394}{$\Delta$}

\def\tsc#1{\csdef{#1}{\textsc{\lowercase{#1}}\xspace}}
\tsc{WGM}
\tsc{QE}
\tsc{EP}
\tsc{PMS}
\tsc{BEC}
\tsc{DE}

\begin{document}
\let\WriteBookmarks\relax
\def\floatpagepagefraction{1}
\def\textpagefraction{.001}

\shorttitle{PINNs for Spatio-Temporal Transformer Ageing Assessment}

\title [mode = title]{Residual-based Attention Physics-informed Neural Networks \\for Spatio-Temporal Ageing Assessment of Transformers \\ Operated in Renewable Power Plants}

\author[1]{Ibai Ramirez}[type=editor,
                        orcid=0009-0005-6581-8641]

\ead{iramirezg@mondragon.edu}

\author[1]{Joel Pino}
\ead{jpino@mondragon.edu}

\author[2,4,8]{David Pardo}
\ead{david.pardo@ehu.eus}

\author[3,4,8]{Mikel Sanz}
\ead{mikel.sanz@ehu.eus}

\author[5]{Luis del Rio}
\ead{lre@ormazabal.com}

\author[5]{Alvaro Ortiz}
\ead{aog@ormazabal.com}

\author[6]{Kateryna Morozovska}
\ead{kmor@kth.com}

\author[7,8]{Jose I. Aizpurua}[type=editor,
                        orcid=0000-0002-8653-6011]
\ead{joxe.aizpurua@ehu.eus}
\cormark[1]

\affiliation[1]{organization={Mondragon University, Electronics \& Computer Science Department},
    addressline={Loramendi, 4.}, 
    city={Mondragon},
    postcode={20500}, 
    country={Spain}}

\affiliation[2]{organization={University of the Basque Country (UPV/EHU),  Department of Mathematics},
    city={Leioa},
    postcode={48080}, 
    country={Spain}}

\affiliation[3]{organization={University of the Basque Country (UPV/EHU),  Department of Physical Chemistry},
    city={Leioa},
    postcode={48080}, 
    country={Spain}}

\affiliation[4]{organization={Basque Centre for Applied Mathematics},
    city={Bilbao},
   postcode={48009}, 
    country={Spain}}
    
\affiliation[5]{organization={Ormazabal Corporate Technology},
    city={Zamudio},
    postcode={48170}, 
    country={Spain}}

\affiliation[6]{organization={KTH, Royal Institute of Technology},
    city={Stockholm},
    postcode={1827}, 
    country={Sweden}}

\affiliation[7]{organization={University of the Basque Country (UPV/EHU),  Department of Computer Science and Artificial Intelligence},
    city={San Sebastian},
    postcode={20018}, 
    country={Spain}}

\affiliation[8]{organization={Ikerbasque, Basque Foundation for Science},
    city={Bilbao},
    postcode={48011}, 
    country={Spain}}

\cortext[cor1]{Corresponding author}

\begin{abstract}
 Transformers are crucial for reliable and efficient power system operations, particularly in supporting the integration of renewable energy. Effective monitoring of transformer health is critical to maintain grid stability and performance. Thermal insulation ageing is a key transformer failure mode, which is generally tracked by monitoring the hotspot temperature (HST). However, HST measurement is complex, costly, and often estimated from indirect measurements. Existing HST models focus on space-agnostic thermal models, providing worst-case HST estimates. This article introduces a spatio-temporal model for transformer winding temperature and ageing estimation, which leverages physics-based partial differential equations (PDEs) with data-driven Neural Networks (NN) in a Physics Informed Neural Networks (PINNs) configuration to improve prediction accuracy and acquire spatio-temporal resolution. The computational accuracy of the PINN model is improved through the implementation of the Residual-Based Attention (PINN-RBA) scheme that accelerates the PINN model convergence. The PINN-RBA model is benchmarked against self-adaptive attention schemes and classical vanilla PINN configurations. For the first time, PINN based oil temperature predictions are used to estimate spatio-temporal transformer winding temperature values, validated through PDE numerical solution and fiber optic sensor measurements. Furthermore, the spatio-temporal transformer ageing model is inferred, which supports transformer health management decision-making. Results are validated with a distribution transformer operating on a floating photovoltaic power plant.
\end{abstract}




\maketitle
\section{Introduction}

The transition towards a decarbonized electric system implies the sustainable and reliable integration of renewable energy sources (RESs) into the power grid. However, this transition faces emerging challenges such as power quality, stability, balance, and power flow issues. These threats result in incorrect equipment operation, accelerated  power equipment ageing, and obstruct the secure, clean, and efficient energy generation, transmission, and distribution process (\citealp{Sinsel_2020}).

In the transition towards the operation of a flexible, decentralized, and resilient power grid, it is necessary to develop advanced functionalities that ensure the reliable and secure delivery of clean energy. In this direction, Prognostics \& Health Management (PHM) focuses on the development of system degradation management solutions through the development of anomaly detection, diagnostics, prognostics, and operation and maintenance planning applications. PHM can be considered as a holistic approach to health management (\citealp{FINK2020103678}).

In the area of PHM, hybrid health monitoring solutions have emerged, which combine physics-based and data-driven models for improved accuracy and robustness (\citealp{Li_2024}). For example, (\citealp{ARIASCHAO2022107961}) focused on remaining useful life (RUL) prediction of a turbofan engine, (\citealp{Lai_24}) addressed anomaly detection of servo-actuators, (\citealp{Shen_EAAI_21}) analysed bearing diagnostics applications, (\citealp{fernandez2023physics}) developed structural element health monitoring applications, and (\citealp{Aizpurua_23}) developed a hybrid transformer RUL approach.

Among hybrid health monitoring solutions, the integration of neural networks (NNs) with partial differential equations (PDEs) through Physics-Informed Neural Network (PINN) approaches (\citealp{wright2022deep, Raisi_19}) have shown promising results, e.g. for complex beam systems (\citealp{Kapoor_PINN_2024}), power systems (\citealp{PowerSystems_23}), or power transformers (\citealp{Bragone22}).

However, due to the PINN model design and training complexity (\citealp{Wang_21, bonfanti2023generalization}), most applications focus on experiments with synthetic data and very few applications use the inspection data to solve practical engineering problems. For example, (\citealp{Kapoor_PINN_2024}) focused on Euler-Bernoulli \& Timoshenko, for complex beam systems;  (\citealp{reyes2021learning}) used Ostwald-de Waele for non-Newtonian fluid viscosity; (\citealp{raissi2020hidden}) used Navier-Stokes for 3D physiologic blood flow and (\citealp{cai2021physics}) employed 2D steady forced convection for heat transfer in electronic chips. The use of PINN models for real operation conditions implies that classical PINN solutions may not be directly applicable. That is, the complexity of real problems (\citealp{pang2019fpinns,kapoor2023physics,kang2022physics}), and convergence issues (\citealp{xiang2022self, Agnostopoulos_RBA}), have been key limitations addressed in the PINN literature. 

Residuals are key PINN parameters that determine the precision and convergence of the PINN model (\citealp{anagnostopoulos2024learning}). However, the stochastic nature of PINNs due to the randomly sampled collocation points results in instability of solutions. Moreover, non-convex loss functions can guarantee a local optimum solution, but there may be other global optima (\citealp{Agnostopoulos_RBA}). With the classical PINN configuration (\citealp{Raisi_19}), residuals at key collocation points may get overlooked and therefore spatio-temporal characteristics may not get fully captured. A solution to this problem is to use weighting multipliers that can be adjusted during training (\citealp{xiang2022self}). An alternative strategy is to use a self-adaptive (SA) attention scheme, which adjusts multipliers through a soft-attention mask (\citealp{MCClenny_SA_2023}) or residual weighting strategies based on residual based attention (RBA) schemes, which have shown promising state-of-the-art results (\citealp{Agnostopoulos_RBA}).

\subsection{Related Work}

Transformers are key integrative assets for the efficient and reliable operation of the grid, and the ever-increasing penetration of RESs to the grid affects the transformer’s health (\citealp{Aizpurua_23}). The main insulating material for oil-immersed transformers is paper immersed in oil and their main failure mode is the insulation degradation. The insulating paper is made of cellulose polymer, in which the number of monomers, also known as the degree of polymerization, determines the strength of the paper (\citealp{IEC60076_transf12}). The insulation paper degradation is the most critical when the temperature of the insulation is the highest, which is known as the winding hotspot temperature (HST). Underestimated HST may lead to reduced transformer cooling operation and the transformer may be running hotter with an accelerated ageing rate. In contrast, overestimated HST may lead to conservative and non-effective maintenance. HST measurements, however, are complex and expensive, and require ad-hoc instrumentation, including fiber-optics and infrared (\citealp{Ping_19}). Even if the transformer is instrumented, the HST is dynamic and the hottest point moves across the windings of the transformer (\citealp{radakovic2003new,deng19}). Accordingly, the accuracy of the HST measurement is reduced.

Data-driven machine learning models have been proposed to accurately estimate the HST, including Support Vector Regression (\citealp{Ruan2021, doolgindachbaporn2021data, deng19}), NN-based transformer thermal models (\citealp{doolgindachbaporn2021data, Aizpurua19}), and probabilistic forecasting models (\citealp{Aizpurua_22,Aizpurua_23}) -- refer to (\citealp{Teh_22_review}) for a comprehensive review. However, it can be observed that existing machine learning based thermal modelling solutions generally focus on the HST and ageing estimation, adopting worst-case temperature estimations and are independent of the spatial location of the HST (refer to Appendix~\ref{appendix:NN_Thermal} for an extended discussion).

Focusing on transformer ageing and lifetime estimation, solutions have been proposed to evaluate ageing through improved feature selection strategies (\citealp{Ramirez_2024_FeatSel}), or even using spectroscopy images and convolutional neural networks (\citealp{Trafo_Ageing_Visual_2024}). However, it can be observed that existing methods evaluate the worst-case HST estimate and use this information to estimate the lifetime of the transformer. In this context, the emergence of PINN models motivates the development of accurate and efficient spatio-temporal transformer ageing models. The main strengths of PINN models with respect to traditional numerical solvers is their capability to (\citealp{Cuomo2022ScientificML}): (i) solve PDEs without meshing the domain, (ii) handle high-dimensional problems more efficiently due to the flexibility of the NN, and (iii) generalize to new conditions or input data without resolving the model. To the best of authors' knowledge, the use of PINNs for improved transformer spatio-temporal ageing estimation has not been addressed before.

Focusing on PINN-based electrical transformer solutions, (\citealp{Bragone22}) presented the first transformer thermal modelling application. This model is based on the following hypotheses (i) uniform heat diffusion in one physical dimension and (ii) steady-state operation conditions. This model estimates the top oil temperature (TOT), tested under stable operation conditions, and it is validated via synthetic finite volume method (FVM) solutions. Published literature shows that existing PINN-based transformer solutions focus on steady-state thermal modelling solutions, with predictable and regular operation profiles. Given the increased penetration of RES, it is of utmost relevance to adapt PINN strategies to evaluate the thermal models that adapt to transient operation states. 

In this context, state-of-the-art PINN solutions for improved PINN accuracy and convergence, such as SA and RBA schemes, have proven to be a very valuable strategy for improved convergence and accuracy (\citealp{Agnostopoulos_RBA}). PINNs have also been used for the ageing assessment of transformers, using inverse configurations to elicit values of degradation-specific parameters (\citealp{Bragone22-ConfAgeing}). However, to the best of authors' knowledge, the use of PINNs for improved spatio-temporal lifetime assessment has not been addressed before.

\subsection{Contribution \& Outline}

This work focuses on the application of PINN methods to infer a spatio-temporal model of the transformer oil and use this information for improved transformer health monitoring strategies. In this direction, the main contributions of this research are the modelling of: (i) the spatio-temporal oil-temperature in transformers operated with RES using efficient and novel attention-based PINN schemes (RBA and SA), (ii) the spatio-temporal transformer winding temperature based on PINN estimates of spatio-temporal oil temperature; and (iii) the spatio-temporal transformer insulation ageing assessment from PINN oil temperature estimates.

The validity of the proposed approach is tested using real inspection data collected at a floating photovoltaic power plant located in Cáceres (Spain). The proposed approach has a direct impact on improved transformer health management, as it enables the efficient and adaptive simulation and generalization of PDEs for their use in transformer thermal modelling and lifetime assessment in renewable power plants.

The remainder of this article is organised as follows: Section~\ref{sec:Problem} defines the transformer thermal and lifetime assessment methods employed in this work. Section~\ref{sec:Approach} introduces the proposed spatio-temporal ageing approach. Section~\ref{sec:CaseStudy} describes the case study. Section~\ref{sec:NumericalResults} applies the proposed approach to the case study, and discusses the numerical results and limitations. Finally, Section~\ref{sec:Conclusions} concludes.

\section{Transformer Thermal Modelling and Lifetime Assessment}
\label{sec:Problem}

\subsection{Thermal modelling using analytic models}
\label{ss:AnalyticIEC}

HST measurements are challenging and expensive. Generally, the HST value, {\small$\Theta_{H}(t)$}, is estimated indirectly from top-oil temperature (TOT) measurements, {\small$\Theta_{TO}(t)$}, as defined in (\citealp{IEC60076_transf12}):

\small
\begin{equation}
\label{eq:HST}
	\Theta_{H}(t)=\Theta_{TO}(t)+\Delta\Theta_{H}(t)
\end{equation}
\normalsize

\vspace{2 mm}
\noindent where $t$\hspace{0.4mm}$\in$\hspace{0.4mm}$\mathbb{R}$ is the time and {\small$\Delta\Theta_{H}(t)$} is the HST rise over TOT, which is given by:

\small
\begin{equation}
	\label{eq:delta_HST}
	\Delta\Theta_{H}(t)=\Delta\Theta_{H_1}(t)-\Delta\Theta_{H_2}(t).
\end{equation}
\normalsize

\vspace{2 mm}
In the above, $\Delta\Theta_{H_1}(t)$ and $\Delta\Theta_{H_2}(t)$ model the oil heating considering the HST variations defined as follows (\citealp{IEC60076_transf12}):

\small
\begin{equation}
        \label{eq:HST_Transient1_1}
         D\Delta\Theta_{H_i}(t)=\upsilon_i\left[\beta_iK(t)^y-\Delta\Theta_{H_i}(t)\right]
\end{equation}
\normalsize

\vspace{2 mm}
\noindent where {\small$K(t)$} is the load factor {\small[p.u.]}, $y$ is the winding exponent constant, which models the loading exponential power with the heating of the windings, {\small$i$\hspace{0.3mm}=\hspace{0.3mm}\{1,\hspace{0.4mm}2\}}, {\small$\upsilon_1$\hspace{0.2mm}=\hspace{0.3mm}$\Delta t/k_{22}\tau_w$}, {\small$\beta_1$\hspace{0.2mm}=\hspace{0.3mm}$k_{21}\Delta\Theta_{H,R}$} both for {\small$i$\hspace{0.3mm}=\hspace{0.3mm}$1$}, and {\small$\upsilon_2$\hspace{0.2mm}=\hspace{0.3mm}$k_{22}\Delta t/\tau_{TO}$}, {\small$\beta_2$\hspace{0.2mm}=\hspace{0.3mm}$(k_{21}$\hspace{0.3mm}-\hspace{0.3mm}$1)\Delta\Theta_{H,R}$} both for {\small$i$\hspace{0.3mm}=\hspace{0.3mm}$2$}. {\small$\Delta t$}\hspace{0.3mm}{\small=}\hspace{0.3mm}$t$\hspace{0.5mm}-\hspace{0.5mm}$t${\small$^\prime$}, {\small$\tau_w$} and {\small$\tau_{TO}$} are the winding and oil time constants, {\small$k_{21}$} and {\small$k_{22}$} are the transformer thermal constants, and {\small$\Delta\Theta_{H,R}$} is the HST rise at rated load. The operator {\small$D$} denotes a difference operation on $\Delta t$, such that {\small$D\Delta\Theta_{H_i}(t)$\hspace{0.3mm}=\hspace{0.3mm}$\Delta\Theta_{H_i}(t)$\hspace{0.3mm}-\hspace{0.3mm}$\Delta\Theta_{H_i}(t^\prime)$} also for {\small$i$\hspace{0.3mm}=\hspace{0.3mm}\{1,\hspace{0.4mm}2\}}. To guarantee numerical stability, {\small$\Delta t$} should be small, and never greater than half of the smaller time constant. 

Under steady state, the initial condition, {\small$\Theta_{H}(0)$}, can be defined as:

\small
\begin{equation}
	\label{eq:InitCond_HST}
        \Theta_{H}(0)=\Theta_{TO}(0)\!+\!k_{21}\Delta\Theta_{H,R}K(0)^y\!-\!(k_{21}\!-\!1)\Delta\Theta_{H,R}K(0)^y
\end{equation}
\normalsize

\vspace{2 mm}
Eq.~(\ref{eq:InitCond_HST}) allows iteratively estimating the next HST values, {\small$\Theta_{H}(n\Delta t)$}, $n\in\mathbb{Z}^+$, using Eqs.~(\ref{eq:HST}), (\ref{eq:delta_HST}), and (\ref{eq:HST_Transient1_1}).

\subsection{Thermal modelling using PDE-based models}
\label{ss:PDE}

Classical transformer thermal models, as defined above, estimate transformer TOT and HST without considering the spatial evolution of the temperature (\citealp{IEEE_Std_C57140, IEC60076_transf12}). This leads to the definition of space-agnostic transformer oil and winding temperature, which can lead to incorrect lifetime estimates. Accordingly, the spatial temperature distribution of the transformer oil and winding is key for the reliable and cost-effective lifetime management. Inspired by (\citealp{Bragone22}), a thermal modelling approach based on partial differential equations (PDE) is developed. The heat-diffusion PDE is considered to model the temporal and spatial evolution of the transformer oil temperature ({\small$\Theta_{O}(x,t)$}). The general form of the one-dimensional (1D) heat-diffusion equation is defined as follows (\citealp{Bragone22}):

\small
\begin{equation}
\label{eq:PDE_1D_Diffusion}
 k\frac{\partial^2\Theta_{O}(x,t)}{\partial x^2} + q(x,t) = \rho c_p \frac{\partial \Theta_{O}(x,t)}{\partial t} \hspace{2mm}\Rightarrow\hspace{2mm} \frac{\partial^2\Theta_{O}(x,t)}{\partial x^2} + \frac{1}{k}q(x,t) = \frac{1}{\alpha} \frac{\partial \Theta_{O}(x,t)}{\partial t}
\end{equation}
\normalsize

\vspace{2 mm}
\noindent where $x,t$\hspace{1.0 mm}$\in$\hspace{0.4mm}$\mathbb{R}$ are the independent variables, which denote position [m] and time [s], respectively, {\small$\Theta_{O}(x,t)$} is given in Kelvin [K], $k$ is the thermal conductivity [W/m.K], $c_p$ is the specific heat capacity [J/kg.K], $\rho$ is the density [kg/m\textsuperscript{3}], $q${\small$(x,t)$} is the rate of heat generation [W/m\textsuperscript{3}], and $\alpha=\frac{k}{\rho c_p}$ is the thermal diffusivity [m\textsuperscript{2}/s].

\begin{figure}[!htb]
	\centering
        \includegraphics[width=0.46\columnwidth]{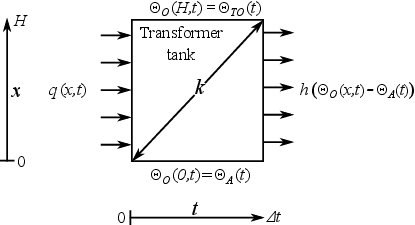}
	\caption{Transformer heat-diffusion model.}
	\label{fig:Trafo_Diffusion_Basic}
\end{figure}

Figure~\ref{fig:Trafo_Diffusion_Basic} shows the thermal parameters of the transformer of the heat-diffusion equation model, which considers the heat source, $q${\small$(x,t)$}, and the convective heat transfer, $h(${\small$\Theta_{O}(x,t)-\Theta_{A}(t)$}$)$.

\vspace{-2 mm}
The heat-source evolution in space and time, $q (x,t)$, is defined as follows:

\small
\begin{equation}
\label{eq:heat-source}
    q(x,t)=P_0+P_K(t)-h\left(\Theta_{O}(x,t)-\Theta_{A}(t)\right)
\end{equation}
\normalsize

\vspace{2 mm}
\noindent where {\small$\Theta_{A}(t)$} is the ambient temperature, $h$ is the convective heat-transfer coefficient, {\small$P_0$} denotes the no-load losses [W], and {\small$P_K(t)$} represents the load losses, defined as follows:

\small
\begin{equation}
    P_K(t)=K(t)^2\mu,
\end{equation}
\normalsize

\vspace{2 mm}
\noindent where {\small$K(t)$} is the load factor [p.u.], and $\mu$ is the rated load losses [W].

The key challenge is to accurately model the transformer oil temperature vertically along its height H, as shown in Figure~\ref{fig:Trafo_Diffusion_Basic}. It can be observed that the spatial distribution is considered along the vertical axis ({\small$x$}) of the transformer. It is assumed that {\small$\Theta_{O}(x,t)$} is equal to: (a) {\small$\Theta_{A}(t)$} at the bottom ({\small$x$\hspace{0.5mm}=\hspace{0.5mm}$0$}), and (b){\small$\Theta_{TO}(t)$} at the top ({\small$x$\hspace{0.5mm}=\hspace{0.5mm}$H$}). Namely, these are the Dirichlet boundary conditions of the PDE that is going to be solved:

\small
\begin{equation}
\label{eq:BoundaryConditions}
\begin{split}
    \Theta_{O}(0,t)&=\Theta_{A}(t)\\
    \Theta_{O}(H,t)&=\Theta_{TO}(t)\\
\end{split}
\end{equation}
\normalsize

\subsubsection{PINN Basics}
\label{ss:PINN_Basics}

PINNs were introduced with the goal of encoding PDE-based physics in machine learning models (\citealp{Raisi_19}), taking advantage of the ability of neural network (NN) models to act as universal approximators (\citealp{wright2022deep}). In the context of PINNs, a general PDE can be expressed as (\citealp{Agnostopoulos_RBA}):

\small
\begin{equation}
\label{eq:PDE_generica}
   \mathcal{D}\{u(x,t)\}=f(x,t)
\end{equation}
\normalsize

\vspace{2 mm}
\noindent where $x,t$\hspace{1.0 mm}$\in$\hspace{0.4mm}$\mathbb{R}$ are independent variables, usually position and time, respectively, $u(x,t)$ is the unknown function to be approximated, $\mathcal{D}$ denotes the differential operator, and $f(x,t)$ is a given forcing function that introduces external influences to the system.

\begin{figure}[!htb]
	\centering
        \includegraphics[width=0.44\columnwidth]{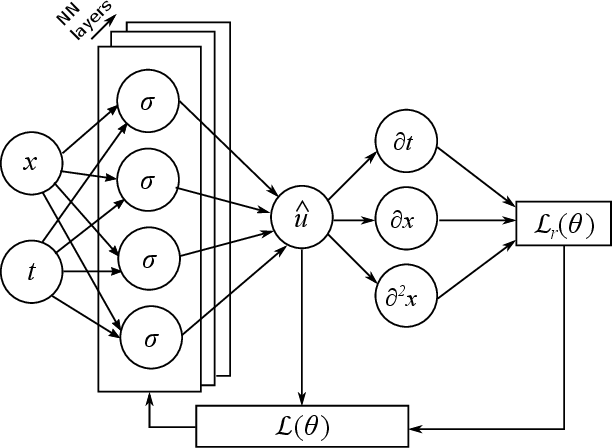}
	\caption{Overall PINN framework.}
	\label{fig:PINN_Framework_General}
\vspace{-5.5 mm}
\end{figure}

 PINNs consist of the NN part, in which the inputs define time ($t$\hspace{0.4mm}$\in$\hspace{0.4mm}$\mathbb{R}$) and space ($x$\hspace{0.4mm}$\in$\hspace{0.4mm}$\mathbb{R}$ for one-dimensional cases) coordinates for the initial conditions (IC) and boundary conditions (BC).The output of the NN is an approximated solution of the PDE at the space and time coordinates, denoted {$\hat{u}(x,t)$}.This is calculated through the iterative application of weights (\bm{$w$}), biases (\bm{$b$}), and non-linear activation functions ($\sigma$) over the input. Namely, the inputs are connected through neurons, where they are multiplied with the weights and summed with the bias term. Finally, the weighted sum is passed through an activation function ($\sigma$). Subsequently, the outcome {$\hat{u}(x,t)$} is post-processed via automatic differentiation (AD) to compute the derivatives in space and time at certain collocation points (CP), generated via random sampling in the interior of the domain. The PINN trains these random points by minimizing the residuals of the underlying PDE, which is the definition of the loss function. Figure~\ref{fig:PINN_Framework_General} synthesizes the overall PINN framework.The loss function ({\small$\mathcal{L}(\bm{\theta})$}) incorporates the prediction error of the NN at IC and BC, and the residual of the PDE estimated via AD at CP:

\vspace{-1 mm}
\small 
\begin{equation}
\label{eq:Loss_PINNgeneral}
\mathcal{L}(\bm{\theta})=\mathcal{L}_0(\bm{\theta})+\mathcal{L}_b(\bm{\theta})+\mathcal{L}_r(\bm{\theta})
\end{equation}
\normalsize

\vspace{2 mm}
\noindent were {\small$\bm{\theta}$\hspace{0.5mm}=\hspace{0.5mm}$\{\bm{w,b}\}$}, while {\small$\mathcal{L}_0(\bm{\theta}),\mathcal{L}_b(\bm{\theta})$} and {\small$\mathcal{L}_r(\bm{\theta}$}) are the loss terms corresponding to IC, BC and the residual of the PDE, respectively:

\vspace{-1 mm}
\small 
\begin{equation}
\label{eq:Loss_IC}
\mathcal{L}_0(\bm{\theta})=\frac{1}{N_{0}}\sum_{i=1}^{N_{0}}|\hat{u}(x_i,0)-u(x_i,0)|^2
\end{equation}

\vspace{1 mm}
\begin{equation}
\label{eq:Loss_BC}
\mathcal{L}_b(\bm{\theta})=\frac{1}{N_{b}}\sum_{i=1}^{N_{b}}|\hat{u}(x_i,t_i)-u(x_i,t_i)|^2
\end{equation}

\vspace{1.5 mm}
\begin{equation}
\label{eq:Loss_r} 
\mathcal{L}_r(\bm{\theta})=\frac{1}{N_r}\sum_{i=1}^{N_r}|(r(x_i,t_i)|^2 
\end{equation}
\normalsize

\vspace{2 mm}
\noindent were {\small$N_0,N_b$} and {\small$N_r$} are the number of IC, BC and residue points, respectively, while {\small$u(x_i,t_i)$} and {\small$r(x_i,t_i)$} denote the known solution and the residual of PDE, respectively, for each training point $i$ defined at the coordinates {\small$(x_i,t_i)$}. According to the PDE defined in Eq.~{(\ref{eq:PDE_generica}) the residual, {\small$r(x,t)$}, is given by the following equation:}

\small
\begin{equation}
\label{eq:Residual_generico}
   r(x,t)=\mathcal{D}\{u(x,t)\}-f(x,t)
\end{equation}
\normalsize

\vspace{2 mm}
Figure~\ref{fig:PINN_Framework_General} synthesizes the overall PINN framework. Note that different from classical NN configurations, which are based on the minimization of the prediction error at the measurement points, PINNs include the effect of the governing physics equations everywhere on the domain.
Minimizing the loss function in Eq.~{(\ref{eq:Loss_PINNgeneral})} using a suitable optimization algorithm provides an optimal set of NN parameters {\small$\bm{\theta}$\hspace{0.3mm}=\hspace{0.3mm}$\{\bm{w,b}\}$}. That is, approximating the PDE becomes equivalent to finding {\small$\bm{\theta}$} values that minimize the loss with a predefined accuracy. Overall, the key training parameters include: number of neurons and number of layers, number of CP, activation function, and the optimizer. Finding the correct solution requires knowing IC and BC. Additionally, the random locations $(x_i,t_i)$, named as CP, are used to evaluate the residual loss [defined in Eq.~(\ref{eq:Loss_r})]. 

The main motivation for using PINNs over numerical methods to solve PDE models is the computational effort and adaptability to different solutions. Namely, PDE models require a mesh of parameters to model and evaluate the PDE. As for PINNs, there is no need to define the whole mesh.

\subsubsection{Transformer Thermal Modelling via PINNs}
\label{ss:DataDriven_HeatDiffusion}

By denoting the estimated transformer oil temperature as {\small$\hat{\Theta}_{O}(x,t)$} the PDE in Eq.~(\ref{eq:PDE_1D_Diffusion}) can be expressed as follows:

\small
\begin{equation}
\label{eq:1D_U_simple}
   \frac{\partial^2\hat{\Theta}_{O}(x,t)}{\partial x^2} + \frac{1}{k}q(x,t)=\frac{1}{\alpha} \frac{\partial \hat{\Theta}_{O}(x,t)}{\partial t}.
\end{equation}
\normalsize

\vspace{1.5 mm}
Similarly, using the above notation and also denoting the estimated ambient temperature and load losses as {\small$\hat{\Theta}_{A}(x,t)$} and {\small$\hat{P}_{K}(x,t)$}, respectively,  the heat-source [cf. Eq.~(\ref{eq:heat-source})] can be expressed as follows:   

\small
\begin{equation}
\label{eq:heat-source_simple}
   q(x,t)=P_0+\hat{P}_K(x,t)-h\left(\hat{\Theta}_{O}(x,t)-\hat{\Theta}_{A}(x,t)\right)
\end{equation}
\normalsize

\vspace{1.5 mm}
Then, the following equation is obtained:

\small
\begin{equation}
\label{eq:1D_simple}
   \frac{1}{\alpha} \frac{\partial \hat{\Theta}_{O}(x,t)}{\partial t}-\frac{\partial^2\hat{\Theta}_{O}(x,t)}{\partial x^2}- \frac{1}{k}\left[P_0+\hat{P}_K(x,t)-h\left(\hat{\Theta}_{O}(x,t)-\hat{\Theta}_{A}(x,t)\right)\right]=0
\end{equation}
\normalsize

\vspace{1.5 mm}
\noindent where, $x,t$\hspace{1.0 mm}$\in$\hspace{0.4mm}$\mathbb{R}$ are the position and time, respectively. The first step to implement a PINN is to define a residual function, $r(x,t)$, using the heat-diffusion equation presented in Eq.~(\ref{eq:1D_simple}):

\small 
\begin{equation}
\label{eq:loss_function}
   r(x,t)=\frac{1}{\alpha} \frac{\partial \hat{\Theta}_{O}(x,t)}{\partial t}-\frac{\partial^2\hat{\Theta}_{O}(x,t)}{\partial x^2}- \frac{1}{k}\left[P_0+\hat{P}_K(x,t)-h\left(\hat{\Theta}_{O}(x,t)-\hat{\Theta}_{A}(x,t)\right)\right]
\end{equation}
\normalsize

\vspace{1.5 mm}
\noindent where {\small$\hat{\Theta}_{O}(x,t)$}, {\small$\hat{\Theta}_{A}(x,t)$} and {\small$\hat{P}_{K}(x,t)$} are approximated by a NN. 

The residual is a main factor to compute the loss function, which is usually given as mean squared error ({\small MSE}). Depending on the architecture of the PINN, the overall loss function may be constituted of various residual terms, including the residual ($\mathcal{L}_{r}$) and other estimations, such as thermal value estimation error ($\mathcal{L}_{\Theta_{O}}$). In the PINN architecture shown in Figure~\ref{fig:PINN_Framework_General}, $\hat{\text{u}}$ represents uni- or multi-variate estimates. In this work, the PINN is used to estimate {\small$\hat{\Theta}_{O}(x,t)$}, {\small$\hat{\Theta}_{A}(x,t)$}, and {\small$\hat{P}_K(x,t)$}. Then, the compound loss function is defined as a function of ambient temperature error ($\mathcal{L}_{\Theta_{A}}$) and load error ($\mathcal{L}_{P_{K}}$), in addition to ($\mathcal{L}_{\Theta_{O}}$) and ($\mathcal{L}_{r}$) defined immediately above:

\small 
\begin{equation}
\label{eq:Loss_MSE_Bragone}
\mathcal{L}(\bm{w},\lambda_{\Theta_{O}},\lambda_{P_{k}},\lambda_{\Theta_A},\lambda_r)=\mathcal{L}_{\Theta_{O}}(\bm{w},\lambda_{\Theta_{O}}) + \mathcal{L}_{P_K}(\bm{w},\lambda_{P_{k}}) + \mathcal{L}_{\Theta_A}(\bm{w},\lambda_{\Theta_A}) + \mathcal{L}_{r}(\bm{w},\lambda_r)
\end{equation}
\normalsize

\vspace{2 mm}
\noindent where $\bm{w}$ are the weights of the NN, and ${\lambda_{\Theta_O}}$, ${\lambda_{P_K}}$, ${\lambda_{\Theta_A}}$, and ${\lambda_r}$ denote the weights assigned to the oil temperature, load, ambient temperature and residual losses, respectively, defined as follows:

\small 
\begin{equation}
\label{eq:Loss_MSE_k}
\mathcal{L}_{P_K}(\bm{w},\lambda_{P_K})=\lambda_{P_K}\frac{1}{N_{b}}\sum_{i=1}^{N_b}|\hat{P}_K(x_i,t_i)-P_K(t_i)|^2
\end{equation}

\begin{equation}
\label{eq:Loss_MSE_Ta}
\mathcal{L}_{\Theta_{A}}(\bm{w},\lambda_{\Theta_A})=\lambda_{\Theta_A}\frac{1}{N_{b}}\sum_{i=1}^{N_{b}}|\hat{\Theta}_{A}(x_i,t_i)-\Theta_{A}(t_i)|^2
\end{equation}

\vspace{0.8 mm}
\begin{equation}
\label{eq:Loss_MSE_u}
\mathcal{L}_{\Theta_{O}}(\bm{w},\lambda_{\Theta_O})=\lambda_{\Theta_O}\frac{1}{N_{b}}\sum_{i=1}^{N_{b}}|\hat{\Theta}_{O}(x_i,t_i)-\Theta_{TO}(t_i)|^2
\end{equation}

\vspace{0.5 mm}
\begin{equation}
\label{eq:Loss_MSE_f}
\mathcal{L}_r(\bm{w},\lambda_r)=\lambda_r\frac{1}{N_c}\sum_{j=1}^{N_c}|r(x_j,t_j)|^2
\end{equation}
\normalsize

\begin{figure}[!htb]
	\centering
	\includegraphics[width=0.44\columnwidth]{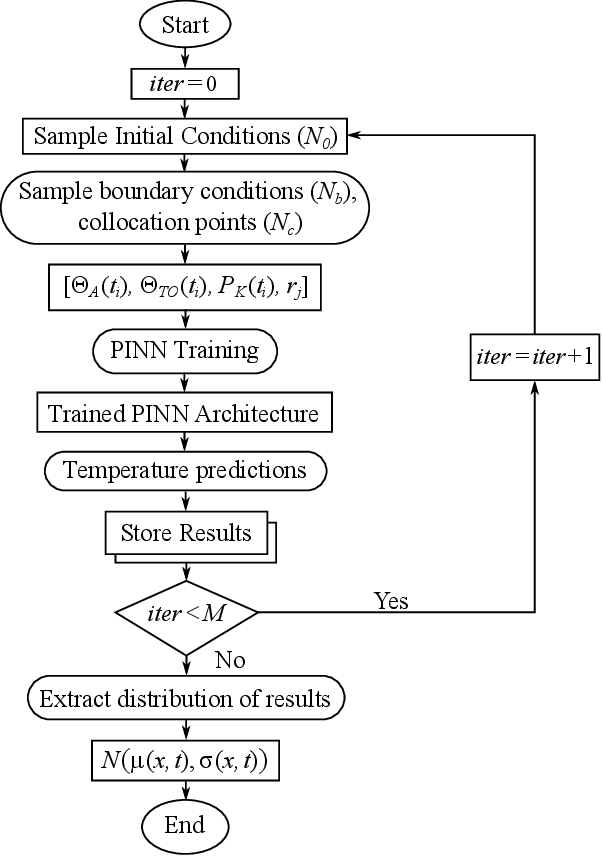}
        \vspace*{-1mm}
	\caption{PINN model-uncertainty assessment framework.}
	\label{fig:PINN_Uncertainty}
        \vspace{-3.5 mm}
\end{figure}

\vspace{0.5 mm}
The points, {\small$\{x_i,t_i,\Theta_{TO}(t_i),P_K(t_i),\Theta_{A}(t_i)\}_{i=1}^{N_b}$} represent the training data given at BC defined in Eq.~(\ref{eq:BoundaryConditions}) and {$\{x_j, t_j\}_{j=1}^{N_c}$} denote the CP for {\small$r(x_j,t_j)$}. $N_b$ is the number of considered training boundary data points, and $N_c$ is the number of collocation data samples {used to compute the residual}. Automatic differentiation is used to infer time- and space- derivatives of the temperature at the selected CP, which are randomly generated using Latin Hypercube Sampling (random number generation in multidimensional space). Note that the definition of the loss function implicitly embeds BC and data term losses, frequently denoted as separate terms in the loss function. Due to the random sampling and stochastic inference of NN models, the PINN has an inherent model-uncertainty, which causes different prediction outcomes. Accordingly, to capture the model-uncertainty in the PINN outcome, the training-testing process is repeated $M$ times, and statistics of the resulting outcome distribution function are reported, specifically, mean and standard deviation. This process evaluates the robustness of the developed PINN model. Figure~\ref{fig:PINN_Uncertainty} shows the overall model-uncertainty assessment process. 

\vspace{-1 mm}

\vspace{0.5 mm}
Namely, in each iteration (\textit{iter}), {\small$N_0$} initial condition points, {\small$N_c$} collocation points and {\small$N_b$} boundary conditions are randomly sampled and the PINN model architecture is adjusted through training. Model training focuses on adapting the NN parameters that minimize the loss in Eq.~(\ref{eq:Loss_MSE_Bragone}).  Subsequently, PINN temperature predictions are inferred and stored from the trained PINN architecture for given input parameters. The process is iteratively repeated $M$ times ($M$ = 100 in this work), and finally, from the stored vector spatio-temporal temperature estimates, assuming a Gaussian distribution, the corresponding mean ($\mu${\small(}\textit{x,t}{\small)}) and variance ($\sigma${\small(}\textit{x,t}{\small)}) values are estimated.

\subsubsection{Self-Adaptive Attention Scheme}
\label{ss:PINNs-SA}

The Self-Adaptive (SA) attention scheme (\citealp{MCClenny_SA_2023}) was inspired from the attention mechanisms used in computer vision, where fully trainable weights are applied to produce a multiplicative soft-attention mask. This scheme updates weights in the loss through gradient descent, which avoid hard-coding them. Using the PDE defined in Eq.~(\ref{eq:1D_U_simple}) and (\ref{eq:heat-source_simple}) and the residual defined in Eq.~(\ref{eq:loss_function}) as reference, the proposed SA-PINN utilizes the following loss function: 
\small 
\begin{equation}
\label{eq:Loss_SA}
\mathcal{L}(\bm{w},\bm{\lambda}_{\Theta_{O}},\bm{\lambda}_{P_{k}},\bm{\lambda}_{\Theta_A},\bm{\lambda}_r)=\mathcal{L}_{\Theta_{O}}(\bm{w},\bm{\lambda}_{\Theta_{O}})+\mathcal{L}_{P_K}(\bm{w},\bm{\lambda}_{P_{k}})+\mathcal{L}_{\Theta_A}(\bm{w},\bm{\lambda}_{\Theta_A}) + \mathcal{L}_{r}(\bm{w},\bm{\lambda}_r)
\end{equation}
\normalsize

\vspace{0.5 mm}
\noindent were $\bm{\lambda}_r${\small\hspace{0.5mm}=\hspace{0.5mm}}$(${\small$\lambda_r^1,\ldots,\lambda_r^{N_c}$}$)$, $\bm{\lambda}_{\Theta_O}${\small\hspace{-0.2mm}=\hspace{0.5mm}}$(${\small$\lambda_{\Theta_{O}}^1,\ldots,\lambda_{\Theta_{O}}^{N_b}$}$)$, $\bm{\lambda}_{P_k}${\small\hspace{-0.2mm}=\hspace{0.5mm}}$(${\small$\lambda_{P_k}^1,\ldots,\lambda_{P_k}^{N_b}$}$)$ and $\bm{\lambda}_{\Theta_A}${\small\hspace{-0.2mm}=\hspace{0.5mm}}$(${\small$\lambda_{\Theta_{A}}^1,\ldots,\lambda_{\Theta_{A}}^{N_b}$}$)$ are trainable, non-negative self-adaptation weights for the residual, boundary and initial points, respectively, and

\vspace{-1mm}
\small 
\begin{equation}
\label{eq:Loss_MSE_k_SA}
\mathcal{L}_{P_K}(\bm{w},\bm{\lambda}_{P_K})=\frac{1}{2}\sum_{i=1}^{N_{b}}m(\lambda_{P_K}^i)|\hat{P}_K(x_i,t_i)-P_K(t_i)|^2
\end{equation}

\vspace{1 mm}
\begin{equation}
\label{eq:Loss_MSE_Ta_SA}
\mathcal{L}_{\Theta_{A}}(\bm{w},\bm{\lambda}_{\Theta_A})=\frac{1}{2}\sum_{i=1}^{N_{b}}m(\lambda_{\Theta_{A}}^i)|\hat{\Theta}_{A}(x_i,t_i)-\Theta_{A}(t_i)|^2
\end{equation}

\vspace{1 mm}
\begin{equation}
\label{eq:Loss_MSE_u_SA}
\mathcal{L}_{\Theta_{O}}(\bm{w},\bm{\lambda}_{\Theta_O})=\frac{1}{2}\sum_{i=1}^{N_{b}}m(\lambda_{\Theta_{O}}^i)|\hat{\Theta}_{O}(x_i,t_i)-\Theta_{TO}(t_i)|^2
\end{equation}

\vspace{0.5 mm}
\begin{equation}
\label{eq:Loss_MSE_f_SA}
\mathcal{L}_r(\bm{w},\bm{\lambda}_r)=\frac{1}{2}\sum_{j=1}^{N_c}m(\lambda_r^i)|r(x_j,t_j)|^2
\end{equation}
\normalsize

\vspace{1 mm}
\noindent where self adaptation mask function $m(\lambda)$ defined on $[0,\infty)$ is a nonnegative differentiable on $[0,+\infty)$ strictly increasing function of $\lambda$ (\citealp{MCClenny_SA_2023}). Gradient ascent is used to update the weight parameters defined as follows:

\vspace{-1.5 mm}
\small
\begin{equation}
\label{eq:Weights_SA}
\lambda_z^{k+1}=\lambda_z^k+\rho_z^k\nabla_{\lambda_z}\mathcal{L}\left(\bm{w},\bm{\lambda}_{P_K},\bm{\lambda}_{\Theta_O},\bm{\lambda}_{\Theta_A},\bm{\lambda}_r\right)
\end{equation}
\normalsize

\vspace{1.4 mm}
\noindent where $\rho_z^k$\hspace{0.8mm}{\small$>$}\hspace{0.3mm}$0$ is a separate learning rate for the self-adaption weights, for $z$\hspace{0.3mm}{\small=}\hspace{0.3mm}\{{\small$P_K,\Theta_O,\Theta_A,$} $r$\}, at step $k$. The gradients, {\small$\nabla_{\lambda_z}\mathcal{L}$}, are computed as follows:

\vspace{-1.0 mm}
\small
\begin{equation}
\label{eq:Gradient_SA}
\nabla_{\lambda_z}\mathcal{L}\left(\bm{w},\bm{\lambda}_{P_K},\bm{\lambda}_{\Theta_O},\bm{\lambda}_{\Theta_A},\bm{\lambda}_r\right)=\frac{1}{2}\begin{bmatrix}m^\prime(\lambda_z^{k,1})|\hat{\Omega}(x_i,t_i)-\Omega(t_i)|^2\\ \ldots \\ m^\prime(\lambda_z^{k,N_b})|\hat{\Omega}(x_i,t_i)-\Omega(t_i)|^2\end{bmatrix}
\end{equation}
\normalsize

\vspace{1.4 mm}
\noindent where {\small$\hat{\Omega}(x_i,t_i)$\hspace{0.3mm}=\hspace{0.3mm}$\{\hat{P}_K(x_i,t_i), \hat{\Theta}_O(x_i,t_i), \hat{\Theta}_A(x_i,t_i)\}$} and {\small${\Omega}(t_i)$\hspace{0.3mm}=\hspace{0.3mm}$\{P_K(t_i),\Theta_O(t_i),\Theta_A(t_i)\}$} for $z$\hspace{0.3mm}{\small=}\hspace{0.3mm}\{{\small$P_K,\Theta_O,\Theta_A$}\}, respectively. The gradient corresponding to the residual, $r$, \textit{i.e.} {\small$\nabla_{\lambda_r}\mathcal{L}$}, is obtained as follows:

\small
\begin{equation}
\label{eq:Gradient_SAr}
\nabla_{\lambda_r}\mathcal{L}\left(\bm{w},\bm{\lambda}_{P_K},\bm{\lambda}_{\Theta_O},\bm{\lambda}_{\Theta_A},\bm{\lambda}_r\right)=\frac{1}{2}\begin{bmatrix}m^\prime(\lambda_r^{k,1})|r(x_i,t_i)|^2\\ \ldots \\ m^\prime(\lambda_r^{k,N_c})|r(x_i,t_i)|^2\end{bmatrix}.
\end{equation}
\normalsize

\vspace{1.4 mm}
\subsubsection{Residual-Based Attention Scheme}
\label{ss:PINNs-RBA}

The Residual-Based Attention scheme (RBA) was introduced to propagate information across the PINN sample space evenly, which is crucial for training PINNs (\citealp{Agnostopoulos_RBA}). The RBA can be seen as a regularization scheme, which adjusts the contribution of individual samples based on the history of preceding samples through an exponentially weighted moving average (\citealp{Agnostopoulos_RBA}):

\begin{equation}
    \lambda_{i}^{t+1}=\gamma \lambda_i^t + \eta^*\frac{|r_i|}{||r_\infty||}
\end{equation}

\vspace{1.5 mm}
\noindent where $i\in\{0,1,\ldots,N_c\}$, is the number of CP, $r_i$ is the residual of each loss term, $\gamma$ is the decay parameter, $\eta^*$ is the learning rate of the weighting scheme, and $\lambda$ refers to the weights used in the loss function as defined in Eq. (\ref{eq:Loss_MSE_Bragone}). 
The adaptation of the weights with cumulative residuals guarantees increased attention on the solution fronts where the PDEs are unsatisfied in spatial and temporal dimensions.

\subsection{Spatio-temporal Winding Temperature and Ageing Assessment}
\label{ss:WindingEstimates}

The transformer thermal model using standard modelling equations estimates the HST with no consideration of their spatial dynamics (see Section~\ref{ss:AnalyticIEC}). In this research, spatio-temporal TOT and HST models are defined, based on the hypothesis of a moving HST across the transformer winding. To this end, a spatio-temporal winding temperature, {\small$\hat{\Theta}_{W}(x,t)$} is defined as follows: 

\vspace{-2mm}
\small
\begin{equation}
\label{eq:HST_spatial}
\hat{\Theta}_{W}(x,t)={\hat{\Theta}_{O}}(x,t)+\Delta\Theta_{H}(t)
\end{equation}
\normalsize

\vspace{1.5mm}
\noindent where, $x,t$\hspace{1.0 mm}$\in$\hspace{0.4mm}$\mathbb{R}$ are the position and time, respectively, {\small$\hat{\Theta}_{O}(x,t)$} is the spatio-temporal oil temperature estimate and {\small$\Delta\Theta_{H}(t)$} is defined in Eq.~(\ref{eq:delta_HST}). In this research, the validity of Eq.~(\ref{eq:HST_spatial}) is tested for winding temperatures at the height of three fibre optic sensors located inside the oil tank. The IEC 60076-7 standard defines insulation paper ageing acceleration factor at  time $t$, {\small$V(t)$}, as (\citealp{IEC60076_transf12}):

\small
\begin{equation}
	\label{eq:ageing_factor}
	V(t)=2^{\left(\Theta_{H}(t)-98\right)/6}
\end{equation}
\normalsize

\vspace{1.0mm}
Based on the spatio-temporal winding temperature estimate in Eq.~(\ref{eq:HST_spatial}), the insulation paper ageing acceleration factor at time $t$ and position $x$, {\small$V(x,t)$}, can be re-defined as:

\small
\begin{equation}
	\label{eq:ageing_factor_spatial}
	V(x,t)=2^{\left(\hat\Theta_W(x,t)-98\right)/6}
\end{equation}
\normalsize

\vspace{1.0mm}
The IEC assumes an expected life of 30 years, with a reference HST of 98$^\circ$C (\citealp{IEC60076_transf12}). The loss of life (LOL) at location $x$ and time $t$ can be defined as:

\small
\begin{equation}
	\label{eq:LoL}
	LOL(x,t)=\int_0^t V(x,t)dt
\end{equation}
\normalsize

\vspace{1.5mm}
Consequently, the LOL at discrete time $L\Delta t$ and location $x$ can be obtained by summing the ageing [cf. Eq.~(\ref{eq:ageing_factor_spatial})] evaluated at the same time instants:

\vspace{-2mm}
\small
\begin{equation}
	\label{eq:lifetime_spatial}
	LOL(x,L\Delta t)=\sum_{n=0}^{L} V(x,n\Delta t)
\end{equation}
\normalsize

\noindent where $n,L$\hspace{1.0 mm}$\in$\hspace{0.4mm}$\mathbb{Z}$\hspace{0.1mm}$^+$.

\vspace{-1.0mm}
\section{Proposed Approach}
\label{sec:Approach}

\begin{figure*}[!htb]
	\centering
	\includegraphics[width=0.84\textwidth]{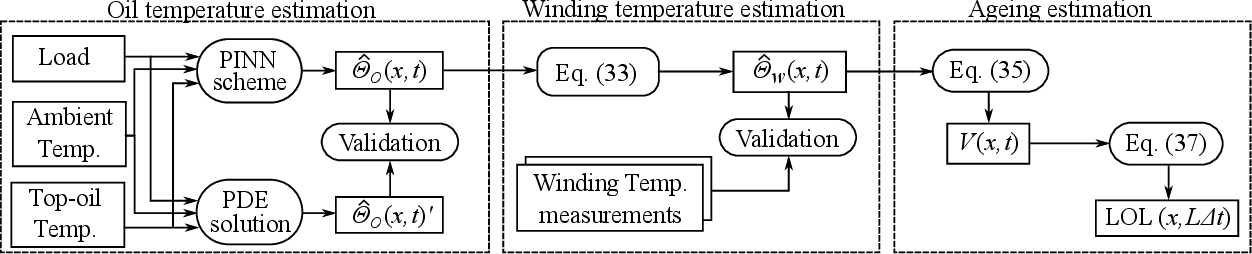}
         \vspace*{-1mm}
         \caption{Overall flowchart of the proposed approach for the efficient spatio-temporal lifetime assessment of transformers.}
	   \label{fig:ApproachFlowchart}
\end{figure*}

Figure~\ref{fig:ApproachFlowchart} shows the proposed spatio-temporal transformer temperature and lifetime estimation approach. The approach starts with the estimation of the transformer oil temperature, {\small$\hat\Theta_O(x,t)$}, which is calculated using the PINN approach (see Figures.~\ref{fig:PINN_Framework_General} and ~\ref{fig:PINN_Uncertainty}) and validated with direct resolution of the PDE. The PINN model can be implemented using different schemes. Three different configurations have been evaluated including the classical vanilla configuration (Vanilla-PINN -- cf. Subsection~\ref{ss:DataDriven_HeatDiffusion}), the self-adaptive attention scheme (SA-PINN -- cf. Subsection~\ref{ss:PINNs-SA}), and the residual-based attention scheme (RBA-PINN -- cf. Subsection~\ref{ss:PINNs-RBA}).

\vspace{-2.0mm}
After obtaining the validated spatio-temporal transformer oil temperature, transformer winding temperature estimates {\small${\hat\Theta}_{W}(x,t)$} are obtained through the application of Eq.~(\ref{eq:HST_spatial}) to the PINN outcome. Results are validated experimentally using winding temperature data collected at different heights using minutely-sampled fiber optic measurements (see Section~\ref{sec:CaseStudy}). 

Finally, the transformer ageing acceleration factor, {\small$V(x,t)$} and the associated spatio-temporal loss of life, \textit{LOL} is calculated through the application of Eq.~(\ref{eq:ageing_factor_spatial}) and Eq.~(\ref{eq:lifetime_spatial}), respectively.

\subsection{Hyperparameter Tuning \& Training Process}
\label{ss:HyperparameterTuning}

Regarding the parameters of the PINN architecture, the BC points, $N_b$, have been tested for several percentages (37.5\%, 50\%, 62.5\%, 75\% and 87.5\%) of the available data (11520 points), \textit{i.e.}, bottom and top oil measurements. The number of CP, $N_c$, is a random sample within the oil tank domain that has been tested for three integer multiples (10, 20, and 40) of the 11520 available measurements. Each combination of parameters was trained and tested 10 times using a 3-hidden layer architecture, where each layer contains 50 neurons. 

Regarding the number of hidden layers and neurons, several PINN architectures were tested as follows: hidden layers = \{2, 3, 4, 5\} and neurons = \{10, 20, 30, 40, 50, 100\}. In this case, each parameter combination was trained and tested 10 times with $N_b$ equal to 75\% of the available data and $N_c$ equal to 10 times the same variable.  

These tests cover a total of 390 different models evaluated for each output. In all experiments, the hyperbolic tangent activation function was used, and training was carried out using the Adam optimizer for 20,000 iterations with a learning rate of 0.001 and L-BFGS-B for a maximum of 50,000 iterations. The predicted temperature values were compared to the resolution outcomes of the PDE model, and the error was quantified through the relative $L_2$ error ($\Gamma_2$), defined as follows:

\vspace{-1.5mm}
\begin{equation}
   \Gamma_2=\frac{||\hat{u}-u||_2}{||u||_2}
\end{equation}

\vspace{1mm}
\noindent where $\hat{u}$ is the predicted value, $u$ is the real measured value, and $||.||_2$ is the $L_2$-norm operator.

\begin{figure*}[!htb]
	\centering
        \includegraphics[width=0.83\textwidth]{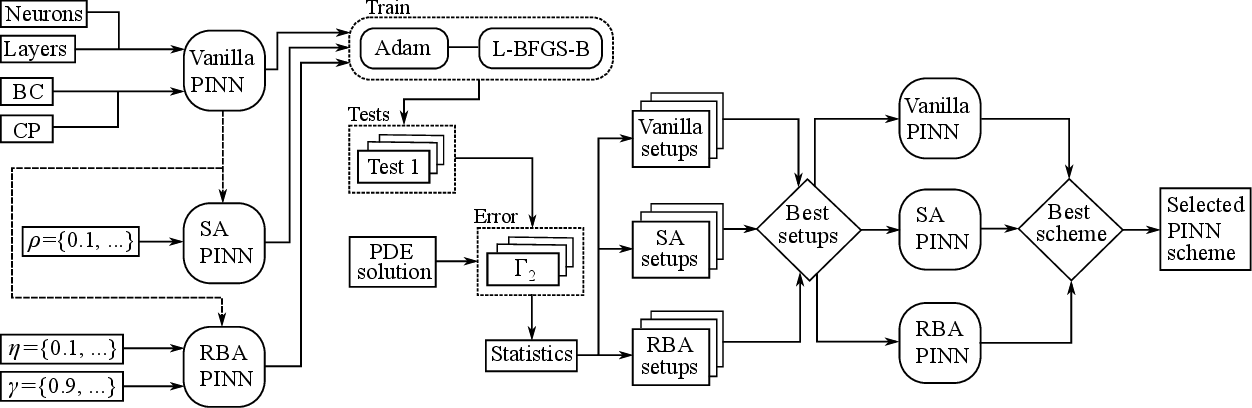}
        \vspace*{-0.2cm}
        \caption{Hyperparameter tuning representation.}
	  \label{fig:HyperTuningFlowchart} 
\end{figure*}

The outlined process has been applied to the main Vanilla-PINN scheme. As for the SA scheme, different learning rate values were tested $\rho_z$\hspace{0.3mm}{\small$=$}\hspace{0.2mm}\{{\small$0.1, 0.01, 0.001$}\} for computing the self-adaption weights, $\lambda_z$, for $z$\hspace{0.3mm}{\small$=$}\hspace{0.3mm}\{{\small$P_K,\Theta_O,\Theta_A,$} $r$\}, which resulted in $\rho_z$\hspace{0.3mm}{\small$=$}\hspace{0.4mm}{\small$0.01$} being selected as it showed the lowest $\Gamma_2$. As for the RBA hyperparameter tuning, a grid of values were tested $\gamma$\hspace{0.3mm}{\small$=$}\hspace{0.2mm}\{{\small$0.9, 0.99, 0.999$}\} and  $\eta$\hspace{0.3mm}{\small$=$}\hspace{0.2mm}\{{\small$0.0001, 0.001, 0.01, 0.1$}\}, resulting in the combination of $\gamma$\hspace{0.2mm}{\small$=$}\hspace{0.3mm}{\small$0.999$} and $\eta$\hspace{0.2mm}{\small$=$}\hspace{0.3mm}{\small$0.001$} being selected as they showed the lowest $\Gamma_2$. Figure~\ref{fig:HyperTuningFlowchart} summarizes the overall hyperparameter tuning, model training and testing processes for the considered PINN schemes.

\vspace{-0.5cm}
\section{Case Study}
\label{sec:CaseStudy}

Sierra Brava is a grid-connected floating PV plant located in the Sierra Brava reservoir (Cáceres, Spain). The installation consists of 5 floating systems, each with 600 PV panels and an estimated capacity of 1.125 MW peak. The focus of the case study is on the lifetime evaluation of the distribution transformer located in the transformation centre (\citealp{Aizpurua_23}). 

\begin{figure}[!htb]
	\centering
        \includegraphics[width=0.54\columnwidth]{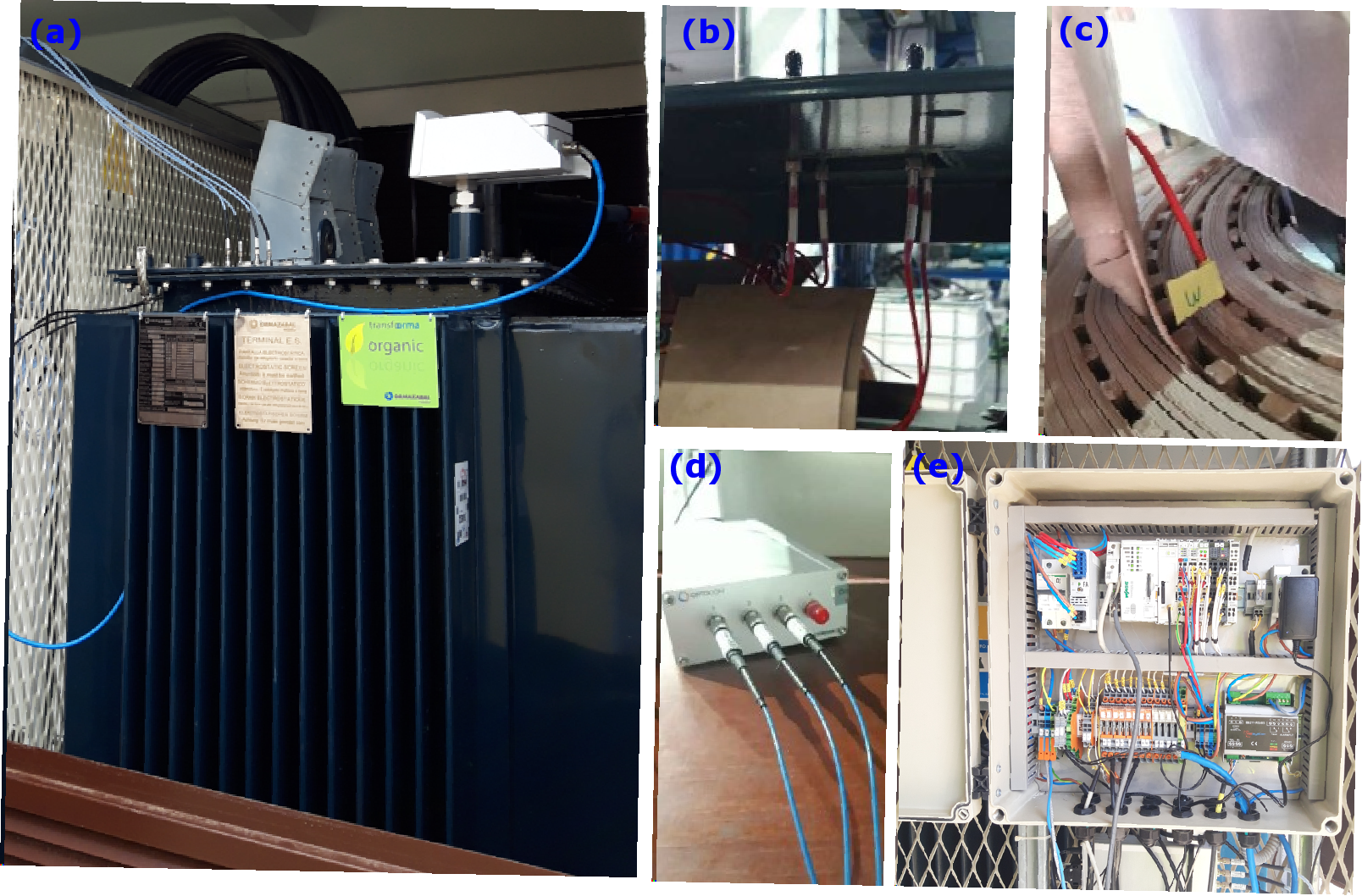}
	\caption{(a) Instrumented transformer, (b) upper side of the transformer, (c) FOS placement in the duct, (d) temperature measurement equipment, and (e) data acquisition and communication setup.}
	\label{fig:Experimental}
\vspace{-0.4cm}
\end{figure}

Namely, three fibre optic sensors (FOS) were installed within the transformer, located at high-voltage (HV) and low-voltage (LV) windings. FOS were deployed inside the middle phase wiring at different winding heights, where, according to expert knowledge, it is expected to locate the HST. Namely, FOS1: HV-LV 70\%; FOS2: HV-LV 85\%; and FOS3: HV 75\%. Figure~\ref{fig:Experimental} shows the experimental setup for transformer winding measurements.

Figure~\ref{fig:Experimental}(b) shows the FOS wiring outputs to the transformer top shield. Figure~\ref{fig:Experimental}(d) shows the measurement equipment, which includes a light source and an optical analyser. Figure~\ref{fig:Experimental}(e) shows the acquisition and communication unit, which uses a standard analogue output and serial communication port for real-time data acquisition. After the installation of FOS, the transformer was ensemble and filled with natural ester oil. Subsequently, heat-run tests were carried out according to IEC 60076-2, as reported in (\citealp{Aizpurua_23}). After this process, thermal parameters $\tau_0$,  $k_{21}$, $k_{22}$ and $\tau_w$ are determined. Table~\ref{table:CaseStudy_trafo} displays the transformer nameplate rating values.

\begin{table}[pos=hbtp]
\centering
\caption{Transformer parameter values.}
\label{table:CaseStudy_trafo}
\begin{tabular}{|>{\centering\arraybackslash}m{5cm}|>{\centering\arraybackslash}m{3.4cm}|} \hline
\small{\textbf{Parameter}} & \small{\textbf{Value}}\\ \hline	
\small{Rating [kVA], V\textsubscript{1}/V\textsubscript{2}} & \small{1100, 22000/400}\\ \hline
\small{R=Load losses/No load losses [W]} & \small{9800/842}\\ \hline		
{\small{$\Delta\Theta_{H,R}$} [$^\circ$C]} & \small{15.1}\\ \hline	
\small{$k_{21}$, $k_{22}$} & \small{2.32, 2.05}\\ \hline	
\small{$\tau_0$, $\tau_w$ [min.]} & \small{266.8, 9.75} \\ \hline		
\end{tabular}
\vspace{-0.2cm}
\end{table}

\section{Results and Discussion}
\label{sec:NumericalResults}

\begin{figure}[!htb]
\vspace{-0.2cm}
	\centering
        \includegraphics[scale=0.48]{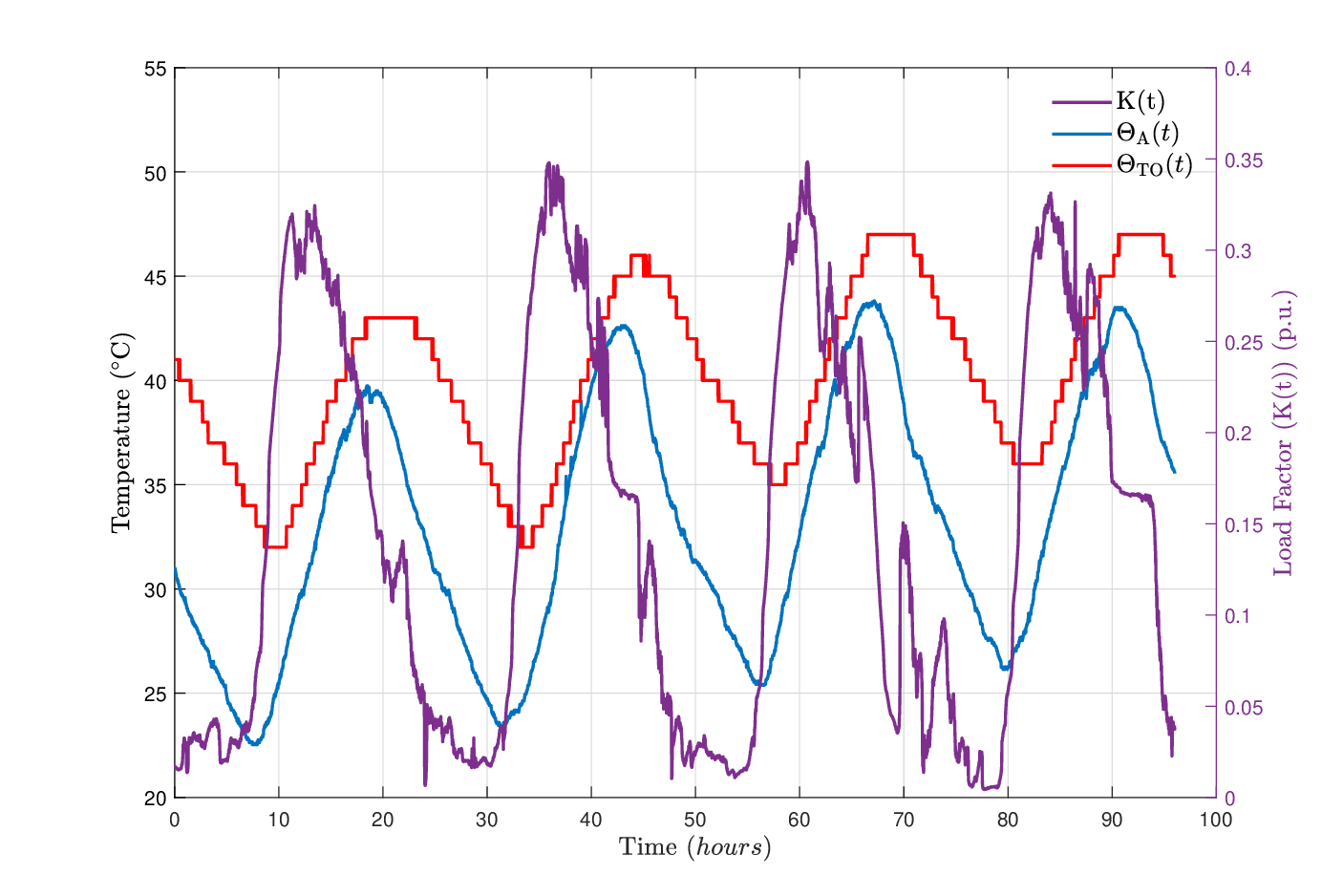}
        \vspace*{-0.2cm}
	\caption{Available load ($K$), ambient temperature ($\Theta_A$) and top-oil temperature ($\Theta_{TO}$), corresponding to 4 days of operation with a sampling rate of 1 minute.}
	\label{fig:AvailableTimeSeries}
 \vspace{-0.3cm}
\end{figure}

Figure~\ref{fig:AvailableTimeSeries} shows the available minutely sampled time-series for ambient temperature, oil temperature,  and load with a total of 5760 samples corresponding to 4 days of operation. Oil and winding temperature profiles analysed throughout the work are focused on minute-based sampling rates. The use of a fast sampling rate compared to oil and winding time constants (Table~\ref{table:CaseStudy_trafo}) allows the evaluation of the transient oil and winding temperature, reflecting the stochastic influence of RESs. 

\begin{figure}[!htb]
\vspace{-0.2cm}
	\centering
        \includegraphics[scale=0.7]{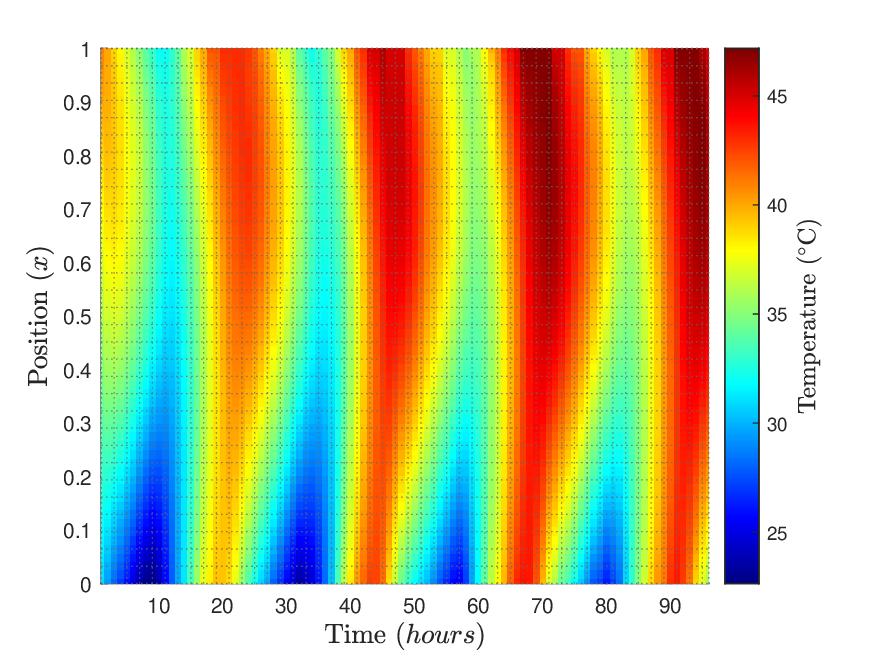}
        \vspace*{-0.2cm}
	\caption{Transformer oil temperature estimation obtained through finite element methods.}
	\label{fig:PDEmatlab}
 \vspace{-0.3cm}
\end{figure}

Figure~\ref{fig:PDEmatlab} shows the numerical solution obtained from the model in Eqs (\ref{eq:PDE_1D_Diffusion})-(\ref{eq:BoundaryConditions}). The PDE has been solved using a finite element method, using MATLAB's \texttt{pdepe} solver, which represents an state-of-the-art solver for this type of PDE (\citealp{matlabpdepe}). It can be observed that the temperature values, {\small$\hat\Theta_{O}(x,t)$}, follow the seasonality of the solar energy, which reflects that the highest power generated by the PV plant occurs at the time of maximum solar irradiation and vice-versa. As for the vertical temperature resolution, it can be observed that the highest temperature values are reached near the top position (\scalebox{0.9}{$x\rightarrow1$}).

\subsection{Oil Temperature Estimation and Validation}
\label{ss:OilTemperature}

\begin{figure}[!htb]
\centering
\includegraphics[width=0.94\textwidth]{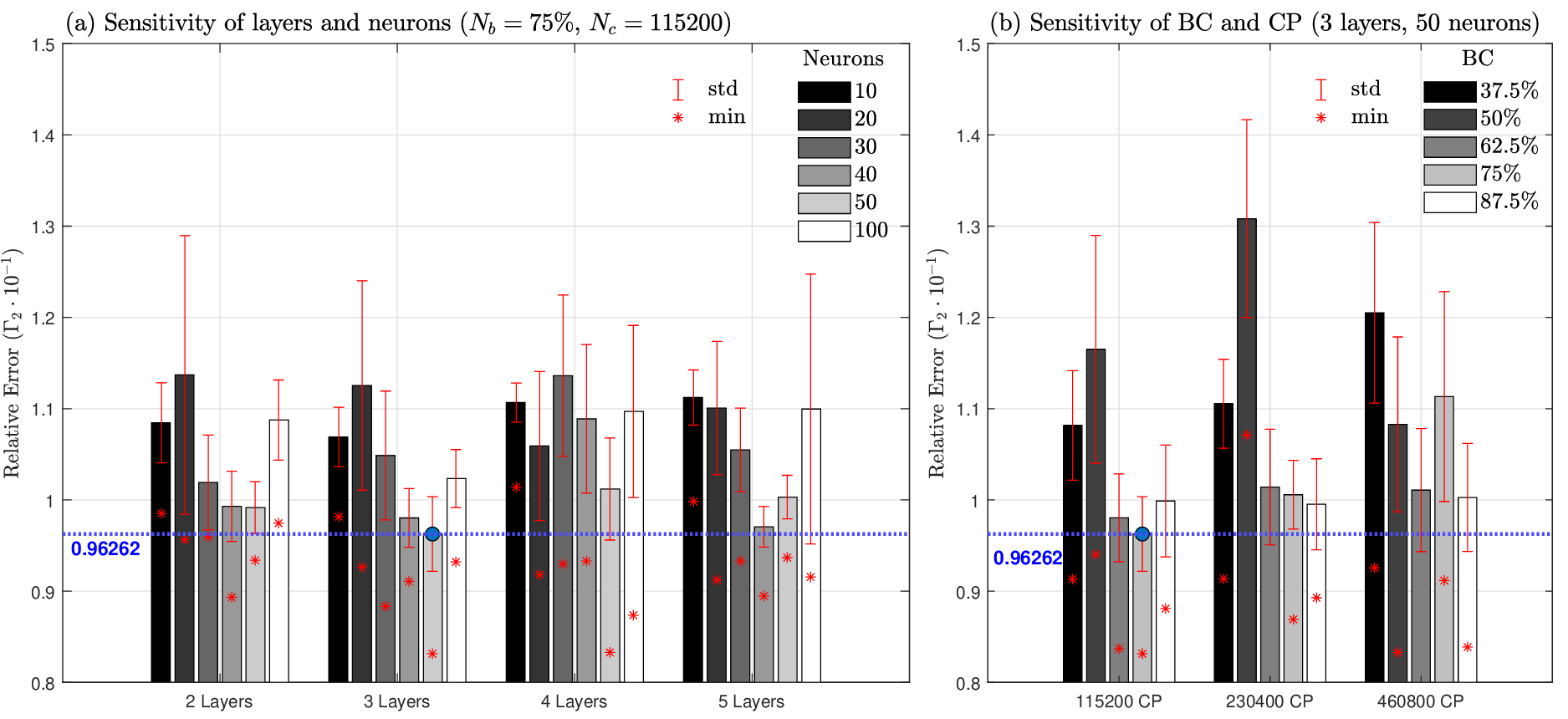}
\vspace*{-0.2cm}
\caption{PINN hyperparameter tuning results for different (a) layers and neurons; and (b) BC and CP.}
\label{fig:PINN_hyperP}
\end{figure}

The transformer oil temperature is usually measured at the top of the tank. However, the oil may have temperature variations across the tank that may be relevant to capture and propagate. Accordingly, in order to design the PINN model, firstly the hyperparameter tuning process defined in Section~\ref{ss:HyperparameterTuning} is adopted. Figure~\ref{fig:PINN_hyperP} shows the obtained results and the best configuration, \textit{i.e.}, 3 layers and 50 neurons, 8640 BC points (75\%), and 115200 CP.

\begin{figure}[!htb]
\vspace{-0.1cm}
	\centering
        \includegraphics[scale=0.64]{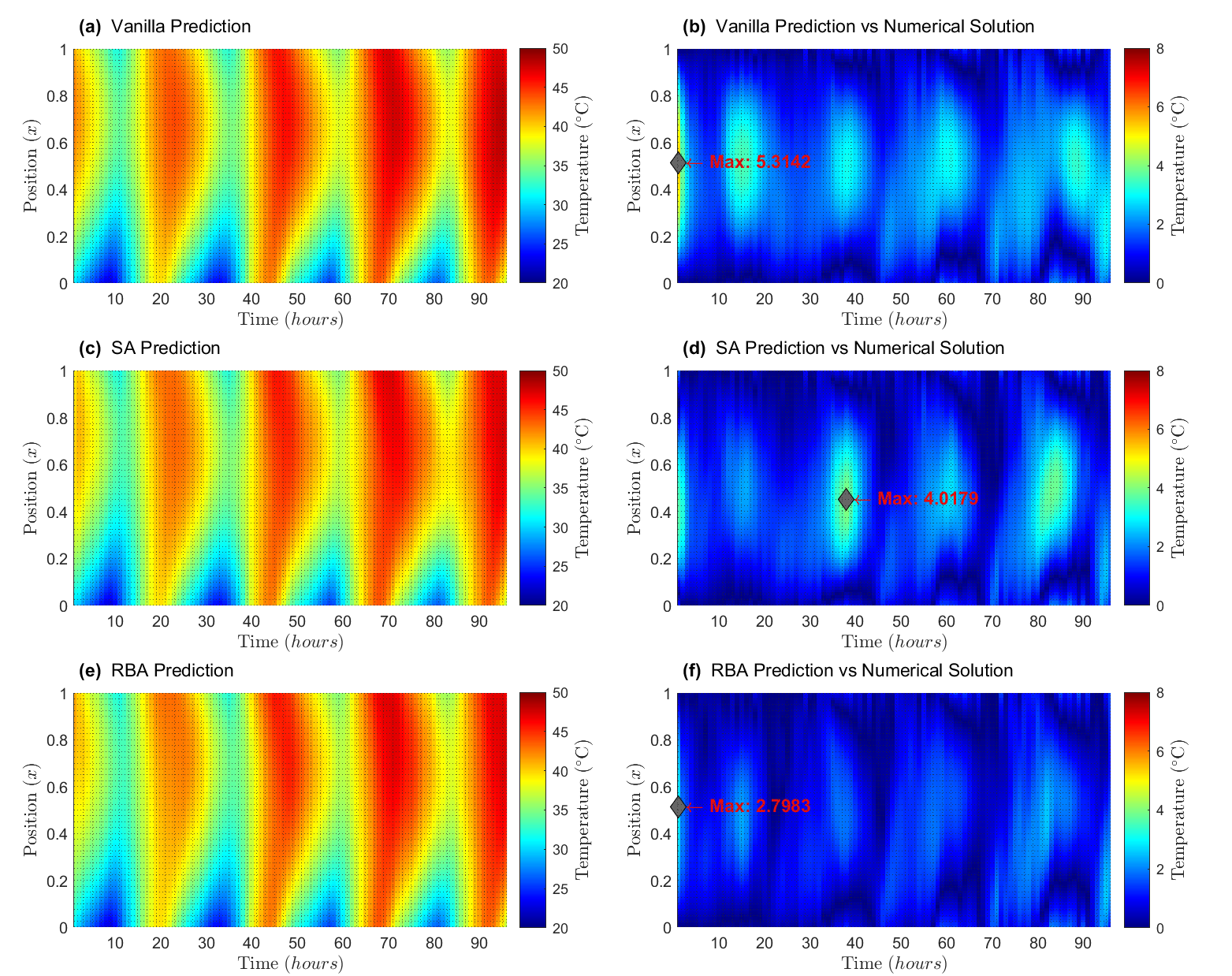} 
        \vspace*{-0.4 cm}
	\caption{Transformer oil temperature predictions for different schemes and error estimates with respect to the numerical solution in Figure \ref{fig:PDEmatlab}. (a) Vanilla-PINN prediction; (b) prediction error of Vanilla-PINN; (c) SA-PINN prediction; and (d) prediction error of SA-PINN; (e) RBA-PINN prediction; (f) prediction error of RBA-PINN.}
	\label{fig:Vanilla_RBA_PINN}
\vspace{-0.3cm}
\end{figure}

Adopting the configuration with the lowest $\Gamma_2$, Figures~\ref{fig:Vanilla_RBA_PINN}(a), \ref{fig:Vanilla_RBA_PINN}(c) and \ref{fig:Vanilla_RBA_PINN}(e) show transformer oil temperature predictions of Vanilla-PINN, SA-PINN and RBA-PINN schemes, respectively. Figures~\ref{fig:Vanilla_RBA_PINN}(b), \ref{fig:Vanilla_RBA_PINN}(d) and \ref{fig:Vanilla_RBA_PINN}(f) show the error of the PINN predictions (Figures~\ref{fig:Vanilla_RBA_PINN}(a), \ref{fig:Vanilla_RBA_PINN}(c) and \ref{fig:Vanilla_RBA_PINN}(e), respectively) with respect to the numerical solution shown in Figure~\ref{fig:PDEmatlab}.

It can be seen from Figures~\ref{fig:Vanilla_RBA_PINN}(e) and \ref{fig:Vanilla_RBA_PINN}(f) that the obtained error values are low. RBA-PINN obtains smaller and more stable errors compared to Vanilla-PINN and SA-PINN models, with a maximum error of 2.8$^\circ$C for RBA-PINN, with respect to a maximum error of 5.3$^\circ$C and 4$^\circ$C, for Vanilla-PINN and SA-PINN, respectively. Therefore, it can be concluded that the developed RBA-PINN model effectively solves the underlying PDE.

\begin{figure}[!htb]
	\centering
        \includegraphics[width=\textwidth]{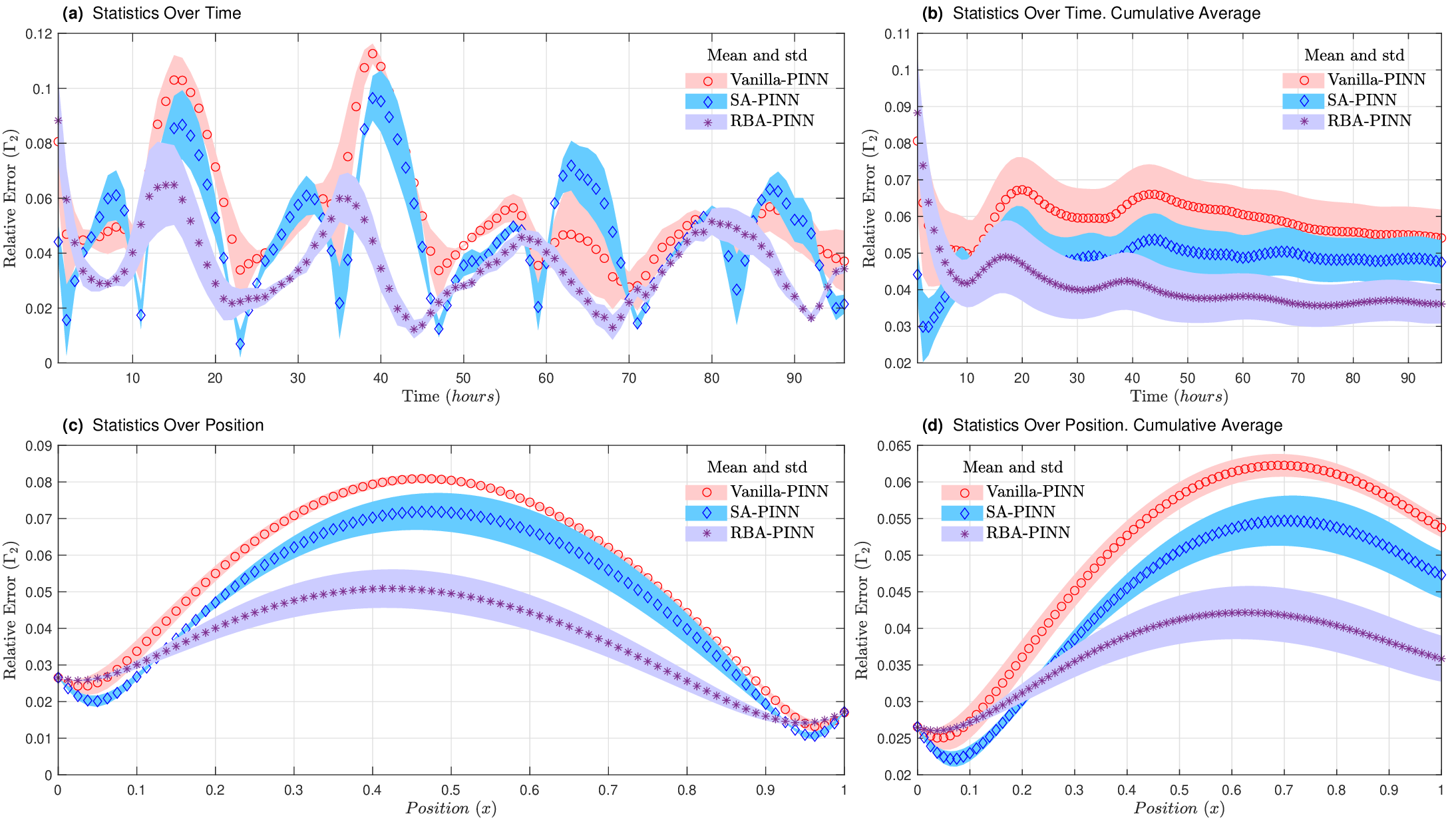}
        \vspace*{-0.4 cm}
	\caption{Relative error of PINN configurations across temporal and spatial dimensions: (a) relative error in time, (b) mean cumulative relative error in time, (c) relative error in space, and (d) mean cumulative relative error in space.}
	\label{fig:Error_Variance}
 \vspace{-0.4cm}
\end{figure}

Figures~\ref{fig:Error_Variance}(a) and \ref{fig:Error_Variance}(b) show the relative error over time and the cumulative average, and Figures~\ref{fig:Error_Variance}(c) and \ref{fig:Error_Variance}(d) show the relative error over spatial dimension $x$ and the cumulative average. It can be observed from Figures~\ref{fig:Error_Variance}(a) and \ref{fig:Error_Variance}(b) that (i) the RBA-PINN configuration has the lowest instantaneous and cumulative relative error across the temporal dimension; (ii) SA-PINN has the best initial conditions among all configurations; and (iii) vanilla-PINN has the highest error among all configurations. Figures~\ref{fig:Error_Variance}(c) and \ref{fig:Error_Variance}(d) show that (i) RBA-PINN has the best performance across the spatial dimension, except near the boundary conditions ($x$\hspace{0.3mm}{\small$=$}\hspace{0.3mm}$0$, $x$\hspace{0.3mm}{\small$=$}\hspace{0.3mm}$1$), where SA-PINN has a better performance; (ii) SA-PINN performs better than Vanilla-PINN configuration for the entire spatial dimension; and (iii) Vanilla-PINN has the narrowest variance among all configurations.

\begin{figure}[!htb]
\vspace{-0.44cm}
	\centering
	\includegraphics[width=0.8\columnwidth]{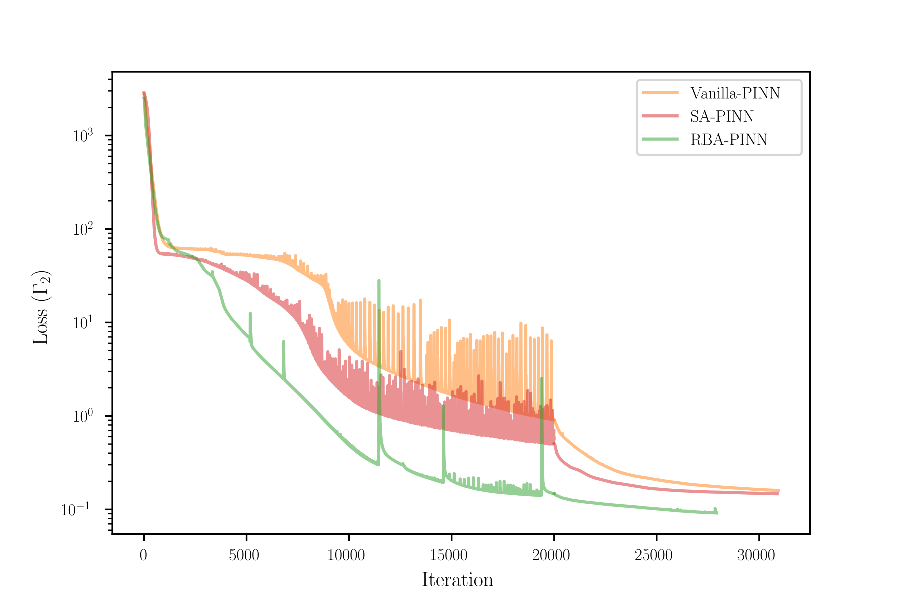}
        \vspace*{-0.3cm}
	\caption{Evolution of the compound loss function of the Vanilla-PINN, RBA-PINN, and SA-PINN models.}
	\label{fig:PINN_Loss_NORBA2RBA}
\end{figure}

Figure~\ref{fig:PINN_Loss_NORBA2RBA} shows the evolution of the loss in Eq.~(\ref{eq:Loss_MSE_Bragone}) as a function of the number of evaluations for the Vanilla-PINN, RBA-PINN and SA-PINN models. It can be seen that the RBA-PINN loss converges faster and obtains loss values smaller than those of the Vanilla-PINN and SA-PINN models. It can also be observed that the Vanilla-PINN and SA-PINN models fluctuate for a longer period of time before reaching a stable loss value. The results of the training process support the PINN outcomes shown in Figures~\ref{fig:Vanilla_RBA_PINN} and \ref{fig:Error_Variance}. 

\begin{figure}[!htb]
\vspace{0.3cm}
	\centering
	\includegraphics[width=0.61\columnwidth]{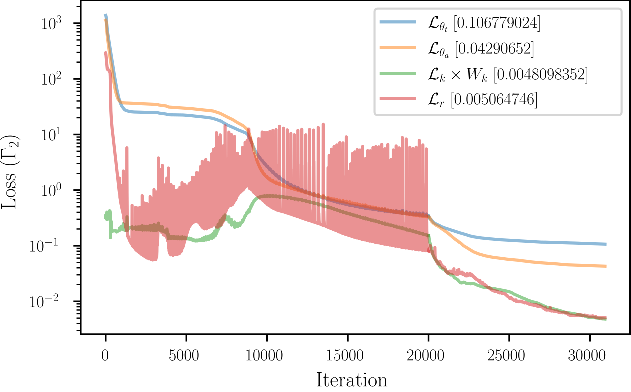}\\
        \includegraphics[width=0.61\textwidth]{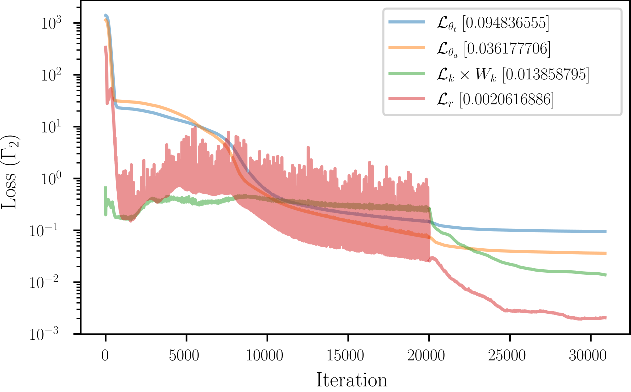}\\
        \includegraphics[width=0.61\textwidth]{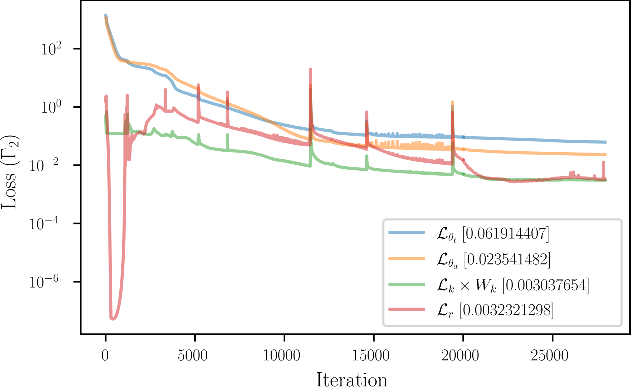}
	\caption{Evolution of the individual compound loss function terms of the Vanilla-PINN (top), SA-PINN (middle), and RBA-PINN (bottom)}.
	\label{fig:PINN_Loss_RBA}
\vspace{-0.2cm}
\end{figure}

Figure~\ref{fig:PINN_Loss_RBA} evaluates the individual loss terms [cf. Eq.~(\ref{eq:Loss_MSE_Bragone})] for each PINN model and it shows that for the Vanilla-PINN and SA-PINN configurations, the MSE of the residual fluctuates through the Adam optimization process (20,000-th iteration). It can be also observed that all the MSE values in the Vanilla-PINN and SA-PINN models are higher than the RBA-PINN model MSE values.

Table~\ref{tab:ComputationCost} displays the computational time for the different PINN configurations for training and evaluation, tested in a computer with AMD Ryzen 9 49000HS, GPU: GeForge RTX 2060 Max-Q, and 32 GB RAM. It can be observed that traditional numerical solvers require substantially less time compared to PINNs if the training stage is taken into account. In contrast, after training the PINN model, the inference with new data is faster. Among PINN configurations, it can be observed that Vanilla-PINN takes the least time, followed by RBA-PINN and SA-PINN models respectively, due to the additional calculations required for the calculation of weighting parameters.

\begin{table}[pos=htb]
    \centering
    \caption{Computation times for different resolution models.}
    \label{tab:ComputationCost}
    \begin{tabular}{|l|c|c|}
        \hline
         \textbf{Method} & \textbf{train [s]} & \textbf{evaluation [s]}  \\ \hline
         Numerical Solver & \multicolumn{2}{c|}{$25.83 \pm 3.69$} \\ \hline
         PINN-Vanilla & $744.62 \pm 62.35$  & $0.62 \pm 0.16$ \\ \hline
         PINN-SA & $2121.02 \pm 134.19$ & $2.58 \pm 0.67$ \\ \hline
         PINN-RBA & $858.50 \pm 163.85$ & $1.50 \pm 0.66$ \\ \hline
    \end{tabular}
\end{table}

\subsection{Winding Temperature Estimation and Validation}
\label{ss:WindingTemperature}

\begin{figure}[!htb]
	\centering
	\includegraphics[width=0.95\columnwidth]{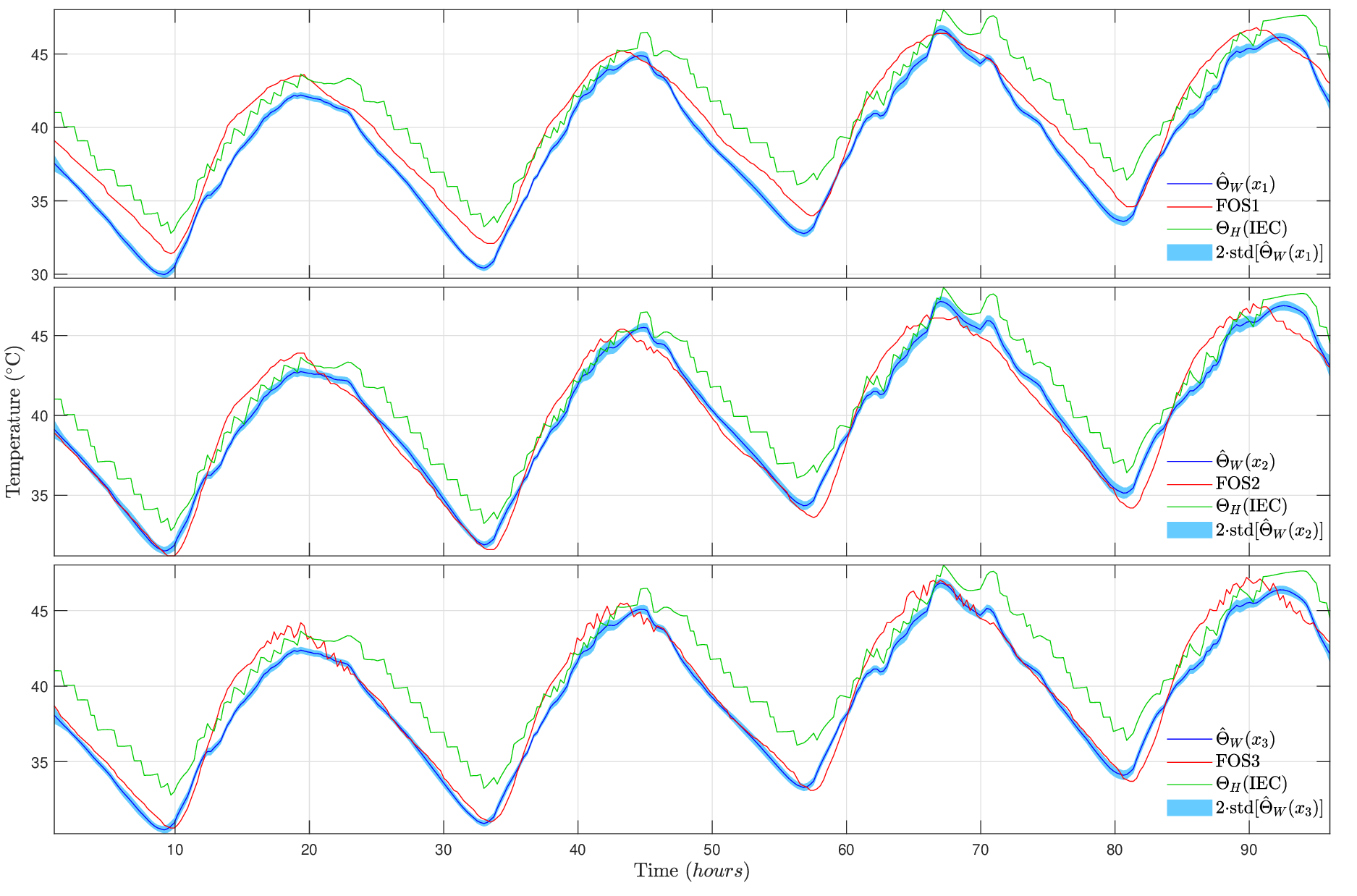}
        \vspace*{-0.3cm}
	\caption{Winding temperature estimation and validation at different heights.}
        \label{fig:Compare_FOS_3D}
\end{figure}

Winding temperature values are inferred from the spatio-temporal oil temperature estimates, as shown in Figure~\ref{fig:ApproachFlowchart} through the most accurate PINN scheme (cf. Figure~\ref{fig:HyperTuningFlowchart}). Figure~\ref{fig:Compare_FOS_3D} shows: (i) three fibre optic sensor-based measurements (FOS1, FOS2, FOS3); (ii) the estimated winding temperatures ({\small$\hat\Theta_{W}(x_1)$}, {\small$\hat\Theta_{W}(x_2)$}, {\small$\hat\Theta_{W}(x_3)$}) at the corresponding heights of the fibre optic sensors ($x_1$=FOS1, $x_2$=FOS2, $x_3$=FOS3) from PINN predictions, using Eq.~(\ref{eq:HST_spatial}); and (iii) the HST value obtained from TOT measurements and through the IEC standard ({\small$\Theta_{H}$(IEC)}) using Eqs.~(\ref{eq:HST})-(\ref{eq:InitCond_HST}). It is observed that the proposed approach accurately estimates the winding temperature values at different heights, in agreement with the FOS measurements. The HST estimation based on IEC equations and TOT measurements (see Section \ref{ss:AnalyticIEC}), now called IEC-HST, does not track the winding temperature at different heights, because IEC-HST is independent of height.
 
\newpage

\begin{figure}[!htb]
\vspace{1mm}
	\centering
        \includegraphics[scale=0.5]{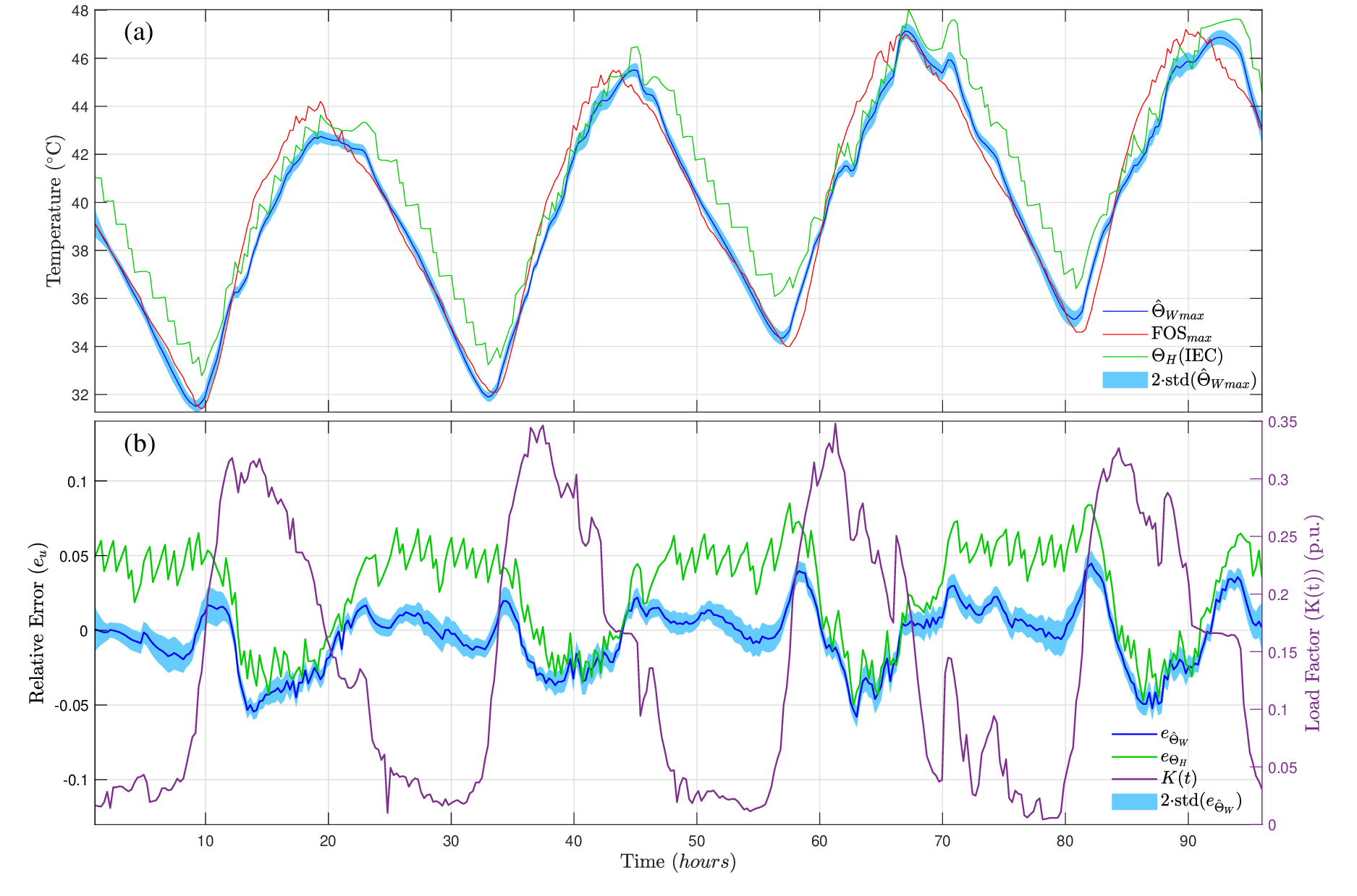}
        \vspace{-4mm}
	\caption{Transformer hotspot temperature (a) estimation and validation; (b) relative error.}
	\label{fig:Winding_Validation}
\end{figure}

Subsequently, the maximum FOS temperature readings are used to estimate the HST from different spatially distributed fibre optic sensors. Similarly, instantaneous maximum PINN-based winding temperature estimates ({\small$\hat{\Theta}_{Wmax}$}) are inferred and compared with the IEC-based HST estimate ({\small$\Theta_{H}$(IEC)}). Figure~\ref{fig:Winding_Validation}(a) shows the obtained HST results and Figure~\ref{fig:Winding_Validation}(b) shows the HST prediction errors, calculated through the relative error, $e_u$, defined as follows: 

\begin{equation}
   e_u=\frac{u-\hat{u}}{u}
\end{equation}

\vspace{1mm}
\noindent where $u$ refers to the ground truth, FOS$\textsubscript{max}$ in this case, and $\hat{u}$ represents the estimates, {\small$\Theta_{H}$(IEC)} and {\small$\hat{\Theta}_{Wmax}$} in this case. Consequently, the errors of the HST estimates based on the IEC and PINN models are denoted by $e${\small$_{\Theta_H}$} and $e${\small$_{\hat\Theta_W}$}, respectively.

\vspace{-2mm}
Figure~\ref{fig:Winding_Validation} shows that IEC-HST slightly underestimates HST during the generation of PV power (daylight) and overestimates during the transformer cooling phase (no PV power, at night). In contrast, the proposed approach ({\small$\hat\Theta_{Wmax}$}) accurately infers the HST. This can be seen in the direct correlation between the PV load and the winding temperature (see Figure~\ref{fig:Winding_Validation}, bottom panel), taking into account the delays in oil and winding heating delays.

\vspace{1mm}
\subsection{Ageing Estimation}
\label{ss:Ageing}

\begin{figure}[!htb]
	\centering
	\includegraphics[width=0.65\columnwidth]{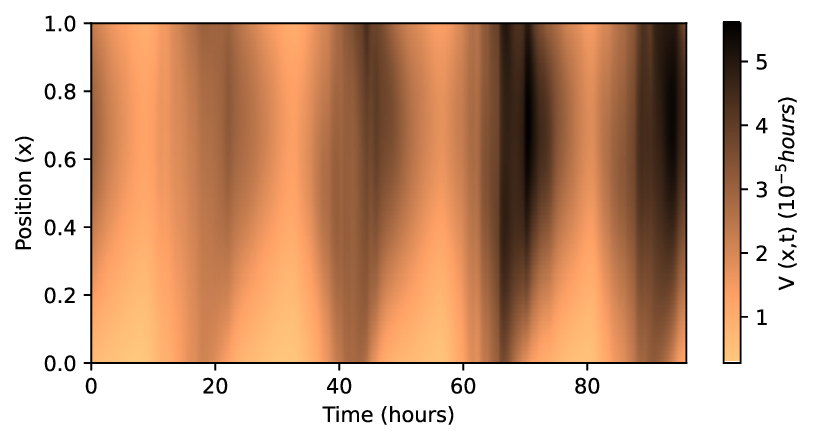} 
        \vspace{-3.5mm}
        \caption{Instantaneous transformer insulation spatial ageing, $V(x,t)$.}
	\label{fig:Spatial_Ageing_Faa}
\end{figure}

Winding temperature estimates have been connected with insulation ageing in Eq.~(\ref{eq:ageing_factor_spatial}) to evaluate the impact of the obtained spatio-temporal winding temperature estimation on ageing. Accordingly, Figure~\ref{fig:Spatial_Ageing_Faa} shows the instantaneous spatio-temporal ageing of the transformer insulation for the considered time-series {(see Figure~\ref{fig:AvailableTimeSeries})} obtained from the estimated winding temperature {(see Figure~\ref{fig:Compare_FOS_3D})}. Namely, each of the spatio-temporal coordinates denotes the ageing inferred in the transformer insulation at location $x$  and instant $t$, $V(x,t)$.

Table~\ref{table:FAA_configs} displays the three key configurations that will be analysed in detail to assess the ageing of the transformer insulation. Namely, configuration \textit{V}{\textit{\scriptsize FOS}} is the ground truth configuration, configuration \textit{V}{\textit{\scriptsize PINN}} is based on the PINN model, which uses the results in Figure~\ref{fig:Spatial_Ageing_Faa}, with the maximum value for each time instant, and configuration \textit{V}{\textit{\scriptsize IEC}} is the IEC-based HST model and associated lifetime estimation.

\begin{table}[pos=h]
\vspace{-2mm}
	\centering
	\caption{Analysed transformer insulation ageing estimation configurations.}
	\label{table:FAA_configs}
        \begin{tabular}{>{\centering\arraybackslash}m{2cm}|>{\centering\arraybackslash}m{6.4cm}|>{\centering\arraybackslash}m{6.2cm}}
            \hline
            \textbf{Configuration} & \textbf{HST estimation} & \textbf{Ageing Model} \\ \hline
            \textit{V}{\textit{\tiny FOS}} & FOS measurements --- Ground Truth & Eq.~(\ref{eq:ageing_factor_spatial}) (using FOS{\small1}, FOS{\small2}, and FOS{\small3}) \\ \hline
            \textit{V}{\textit{\tiny PINN}} & PINN spatio-temporal model -- via Eq.~(\ref{eq:HST_spatial}) & Eq.~(\ref{eq:ageing_factor_spatial}) (using {\small$\hat\Theta_{W}(x,t)$}) \\ \hline
            \textit{V}{\textit{\tiny IEC}} & IEC-HST -- via Eq.~(\ref{eq:HST}) & Eq.~(\ref{eq:ageing_factor}) (using {\small${\Theta}_{H}(\text{IEC})$}) \\ \hline
        \end{tabular}
\vspace{-1mm}
\end{table}

\begin{figure}[!htb]
\centering
\includegraphics[scale=0.48]{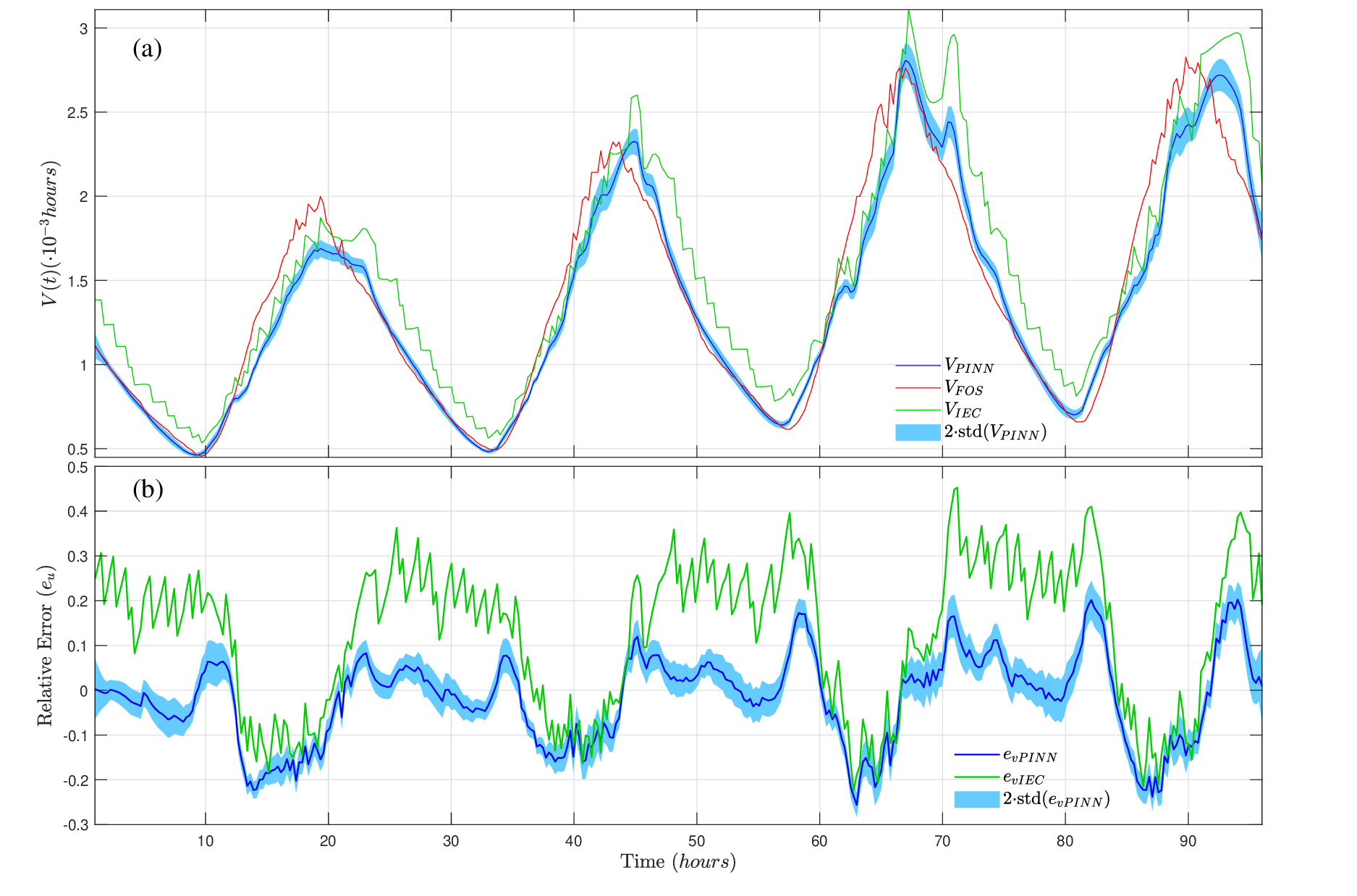}
\vspace{-4mm}
\caption{(a) Instantaneous maximum ageing estimation and validation (b) relative error.}
\label{fig:Spatio_Temporal_Ageing}
\end{figure}

Figure~\ref{fig:Spatio_Temporal_Ageing}(a) shows the instantaneous maximum ageing results for the analysed  configurations. Figure~\ref{fig:Spatio_Temporal_Ageing}(b) shows the errors of the instantaneous maximum ageing based on the PINN (\textit{V}{\textit{\scriptsize PINN}}) and  IEC (\textit{V}{\textit{\scriptsize IEC}}) models, with respect to the ground truth (\textit{V}{\textit{\tiny FOS})}, denoted by $e${\small$\textsubscript{\textit{vPINN}}$} and $e${\small$\textsubscript{\textit{vIEC}}$}, respectively.

It can be seen from Figure~\ref{fig:Spatio_Temporal_Ageing} that configuration \textit{V}{\textit{\tiny PINN}} most accurately tracks \textit{V}{\textit{\tiny FOS}}. It can be also seen that \textit{V}{\textit{\tiny IEC}} overestimates transformer insulation ageing due to the worst-case heating process assumed by the standard. At the end of the analysed period, the maximum cumulative loss-of-life [cf. Eq.~(\ref{eq:lifetime_spatial})] is of 7.65 minutes for the ground truth (\textit{V}{\textit{\tiny FOS}}). At the same instant, the relative error of the RBA-PINN based spatio-temporal ageing (\textit{V}{\textit{\tiny PINN}}) is of 4.2\%, and for IEC-based ageing (\textit{V}{\textit{\tiny IEC}}) is of 9.98\%, with respect to the ground truth.

Comparing the cumulative ageing estimates obtained from the proposed approach and the IEC-HST accumulated LOL results with respect to the ground truth, it is observed that the proposed approach accurately tracks the spatial evolution of the winding temperature (see Figure~\ref{fig:Winding_Validation}) and, accordingly, infers spatio-temporal transformer insulation ageing estimates (see Figure~\ref{fig:Spatio_Temporal_Ageing}). On the contrary, LOL estimates based on IEC-HST do not track winding temperature (see Figure~\ref{fig:Winding_Validation}), and the inferred ageing estimate may be biased (see Figure~\ref{fig:Spatio_Temporal_Ageing}). This will have implications on the transformer O\&M practice, leading to accurate maintenance decisions based on estimated ageing. Note that Figure~\ref{fig:Spatio_Temporal_Ageing} results from the application of the ageing models in Table~\ref{table:FAA_configs} to Figure~\ref{fig:Winding_Validation}. Accordingly, their main difference is on the signal amplitude, which impacts on the corresponding relative error estimate.

\subsection{Limitations}

The final objective of the proposed spatio-temporal transformer insulation ageing assessment approach is to accurately and efficiently estimate the localized ageing in the transformer insulation. This will enable the transformer engineers to accurately plan transformer maintenance actions, adopt safe life-extension decisions, and reduce conservative maintenance decisions.

The results presented in this work show that it is feasible to improve accuracy and computational cost with respect to existing engineering practice based on international standards. However, this research has adopted a simplified 1D heat-diffusion model with uniform heating and without oil convection. This may be a limiting factor because it does not consider additional spatial dimensions. In this direction, the inclusion of additional spatial dimensions will improve the spatial resolution and accuracy. In addition, PINNs are particularly effective for high-dimensional problems (\citealp{Hu_2024}) and the extension into higher dimensions will highlight the computational efficiency of PINNs with respect to classical PDE solvers.

Training PINN models for practical engineering problems is a challenge. The lack of a synthetic ground truth, often given by known PDE solutions, hampers the convergence of the PINN. Measurement uncertainties directly impact on the physics-based part of the loss function and influence the convergence.  In this direction, this research has adopted state-of-the-art PINN training schemes that partly alleviate this problem. However, as the dimension of the problem increases, it is to be expected that the training process will be more complex too.

The adopted PDE determines the validity of the proposed approach. That is, if the physical properties of the thermal insulation change, e.g. through the transformer end of life, the underlying physical model may not capture the thermal properties accurately, and therefore it will have an impact on the PINN solution. In this direction, it may have been possible to improve the PDE model to ensure the accuracy of the physics-based model throughout the transformer ageing.

\section{Conclusions}
\label{sec:Conclusions}

This research presents a novel, accurate, and efficient spatio-temporal transformer winding temperature and insulation ageing model based on Physics Informed Neural Networks (PINNs). The proposed approach leverages physics-based partial differential equations (PDE) with data-driven learning algorithms to improve prediction accuracy and acquire spatial resolution. Transformer oil heating dynamics have been modelled through a spatio-temporal transformer heat diffusion model, defined in a single physical dimension. This model has been coupled with a neural network (NN) architecture to regularize the NN loss function. Training and validation of the PINN has been performed with real inspection data, using different PINN schemes. Namely, the vanilla, self-adaptive (SA), and residual-based attention (RBA) PINN schemes have been implemented.

PINN-based oil temperature predictions are used to estimate spatio-temporal transformer winding temperature values, which are validated using PDE resolution models and fiber-optic sensor measurements, respectively. The spatio-temporal transformer insulation ageing model is also inferred, which supports the transformer health management decision-making and can inform about the effect of localized thermal ageing phenomena in the transformer insulation. The approach is valid for transformers operated in renewable power plants, as it captures fast-changing operation dynamics. The results are validated with a distribution transformer operated on a floating photovoltaic power plant. Results show that the RBA-PINN scheme obtains the most accurate results followed by SA-PINN and vanilla-PINN, with worst-case temperature estimation errors of 2.7$^\circ C$, 4$^\circ C$, and 5.3$^\circ C$ respectively.

Accurately training PINN models is a challenge. In this direction, this research has adopted a simplified 1D heat-diffusion model with uniform heating, without oil convection and state-of-the-art attention mechanisms. Future work may address the use of more detailed PDE models in higher dimensions. However, as the complexity of the problem increases, the underlying necessary PINN model is more complex, the volume of the space increases quickly, and the available data become sparse. In this direction, the increase of the thermal modelling complexity in higher dimensions may lead to explore and adopt recent PINN formalisms, such as hierarchically normalized PINNs (\citealp{LeDuc_24}), which improve training efficiency and solution accuracy. Other alternatives may be to explore more realistic transformer winding topologies through $\Delta$-PINN formalisms (\citealp{DeltaPINN_24}) or to quantify all sources of uncertainty using appropriate methodologies (\citealp{Zhang_Uncertainty_19,Psaros_2023, BayesPINN}).

\section*{CRediT authorship contribution statement}

\textbf{I. Ramirez}: Conceptualization, Methodology, Investigation, Software, Visualization, Data Curation, Writing - Original Draft.
\textbf{J. Pino}: Conceptualization, Methodology, Investigation, Formal Analysis, Validation, Writing - Original Draft.
\textbf{D. Pardo}: Conceptualization, Funding acquisition, Writing - Review \& Editing.
\textbf{M. Sanz}: Conceptualization, Funding acquisition, Writing - Review \& Editing.
\textbf{L. del Rio}: Resources, Funding acquisition, Writing - Review \& Editing.
\textbf{A. Ortiz}: Resources, Writing - Review \& Editing.
\textbf{K. Morozovska}: Writing - Review \& Editing.
\textbf{Jose I. Aizpurua}: Conceptualization, Methodology, Investigation, Validation, Writing - Original Draft, Supervision, Project administration, Funding acquisition.

\section*{Declaration of competing interest}

The authors declare that they have no known competing financial interests or personal relationships that could have appeared to influence the work reported in this paper.

\section*{Acknowledgements}

This research was funded by the Department of Education of the Basque Government (EJ-GV), IKUR Strategy and by the EU NextGenerationEU/PRTR and Spanish State Research Agency (AEI) (grant No. CPP2021-008580). J. I. A. acknowledges financial support (FS) from AEI, Ramón y Cajal Fellowship (grant number RYC2022-037300-I), co-funded by MCIU/AEI/10.13039/501100011033 and FSE+; and from the EJ-GV through the Elkartek (grant number KK-2024/00030) and Consolidated Research Group  (grant No. IT1504-22) programs. KM acknowledges FS from Vinnova (grant No. 2021-03748 \& 2023-00241). DP acknowledges FS from PID2023-146678OB-I00 funded by MICIU/AEI /10.13039/501100011033 and by the EU NextGenerationEU/ PRTR; “BCAM Severo Ochoa” accreditation of excellence CEX2021-001142-S funded by MICIU/AEI/10.13039/501100011033; EJ-GV through the BERC 2022-2025; Elkartek (grant No. KK-2023/00012 \& KK-2024/00086); and Consolidated Research Group MATHMODE (IT1456-22) programs. MS acknowledges FS from HORIZON-CL4-2022-QUANTUM01-SGA project 101113946 OpenSuperQPlus100 EU Flagship on Quantum Tech., AEI grants RYC-2020-030503-I \& PID2021-125823NA-I00 MCIN/AEI/10.13039/501100011033, “ERDF A way of making EU”, “ERDF Invest in your Future”, and EJ-GV (grant No. IT1470-22). The authors would like to thank Iker Lasa at Tecnalia Research \& Innovation for his comments and useful discussions.

\bibliographystyle{model2-names.bst}


\bibliography{myBib}

\newpage

\appendix
\section{Neural Network based Transformer Thermal Predictions}
\label{appendix:NN_Thermal}

Classical Neural Network (NN) models have been tested to evaluate the difference with respect to the PINN model. NN model training is based on purely data-driven loss function, which does not include any physics information.

\begin{figure}[h]
	\centering
        \includegraphics[scale=0.4]{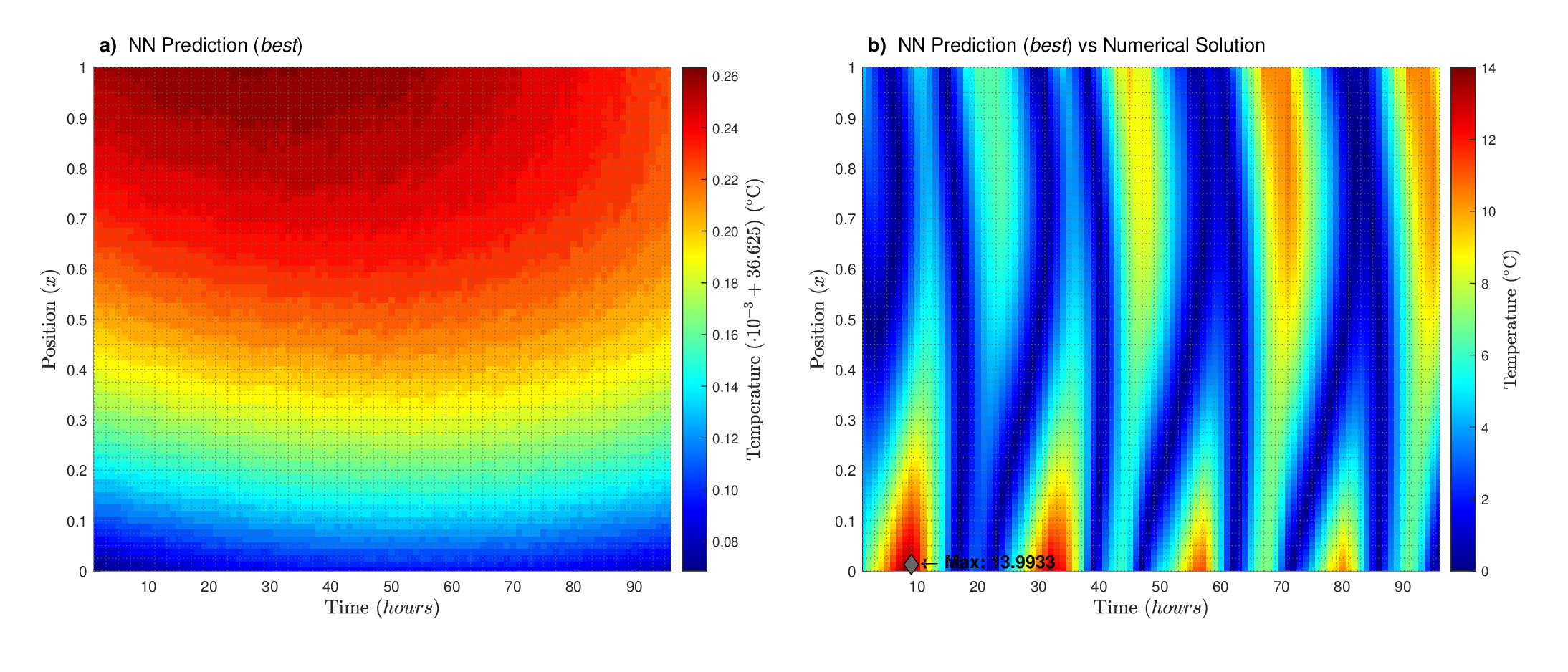}
	\vspace*{-.5 cm}
        \caption{Best prediction results for the NN model: (a) oil temperature prediction; (b) error with respect to the PDE solution in Figure~\ref{fig:PDEmatlab}.}
	\label{fig:NN_best}
\end{figure}

In this case, a number of NN configurations were trained varying the number of hidden layers and neurons (Adam optimizer with 20000 iterations), and best configuration results are reported (NN with 3 layers and 50 neurons). NN model simulations are repeated $70$ times, as with the PINN model (cf. Figure~\ref{fig:ApproachFlowchart}), and Figure \ref{fig:NN_best} shows the best result (lowest error) of the NN predictions compared with the PDE resolution in Figure~\ref{fig:PDEmatlab}.

From Figure \ref{fig:NN_best}(a) it is observed that the results lack of physical meaning and this is reflected in the prediction error results in Figure \ref{fig:NN_best}(b).

\end{document}